\documentclass[journal]{IEEEtran}

\usepackage{xcolor,soul,framed} 

\colorlet{shadecolor}{yellow}
\usepackage[pdftex]{graphicx}
\graphicspath{{../pdf/}{../jpeg/}}
\DeclareGraphicsExtensions{.pdf,.jpeg,.png}

\usepackage[cmex10]{amsmath}
\usepackage{array}
\usepackage{mdwmath}
\usepackage{mdwtab}
\usepackage{eqparbox}
\usepackage{url}
\usepackage{cite}
\usepackage{booktabs}

\begin{document}
\bstctlcite{IEEEexample:BSTcontrol}
    \title{Deep Learning in Palmprint Recognition-A Comprehensive Survey}
  \author{Chengrui~Gao,~\IEEEmembership{Student Member,~IEEE,}
      Ziyuan~Yang, Wei Jia,~\IEEEmembership{Member,~IEEE,} Lu Leng,~\IEEEmembership{Member,~IEEE,} Bob~Zhang,~\IEEEmembership{Senior Member,~IEEE,}
    and Andrew Beng Jin Teoh,~\IEEEmembership{Senior Member,~IEEE}
    \thanks{\textit{This work has been submitted to the IEEE for possible publication. Copyright may be transferred without notice, after which this version may no longer be accessible.}}
\thanks{This work was supported in part by the National Research Foundation of Korea (NRF) under Grant NRF-2022R1A2C1010710; in part by the Science and Technology Development Fund, Macao S.A.R~(FDCT) 0028/2023/RIA1.(Corresponding author: Ziyuan Yang and Andrew Beng Jin Teoh)}
  \thanks{Chengrui Gao is with the College of Computer Science, Sichuan University, Chengdu 610065, China (e-mail: chengrui0101@gmail.com).}
  \thanks{Ziyuan Yang is with the College of Computer Science, Sichuan University, Chengdu 610065, China and also with Centre for Frontier AI Research~(CFAR), Agency for Science, Technology and Research~(A*STAR), Singapore (e-mail: cziyuanyang@gmail.com).}%
  \thanks{Wei Jia is with the School of Computer and Information, Hefei University
of Technology, Hefei, China (e-mail: jiawei@hfut.edu.cn).}
\thanks{Lu Leng is with the Jiangxi Provincial Key Laboratory of Image Processing and Pattern Recognition, Nanchang Hangkong University, Nanchang, China (e-mail: leng@nchu.edu.cn).}
  \thanks{Bob Zhang is with the Pattern Analysis and Machine Intelligence Group, Department of Computer and Information Science, University of Macau, Taipa, Macau, China (e-mail: bobzhang@um.edu.mo).}
  \thanks{Andrew Beng Jin Teoh is with the School of Electrical and Electronic Engineering, College of Engineering, Yonsei University, Seoul, Republic of Korea (e-mail: bjteoh@yonsei.ac.kr).} 
  }

\markboth{Journal of \LaTeX\ Class Files,~Vol.~14, No.~8, August~2015}{Gao \MakeLowercase{\textit{\textit{et al}.}}: High-Efficiency Diode and Transistor Rectifiers}

\maketitle

\begin{abstract}

Palmprint recognition has emerged as a prominent biometric technology, widely applied in diverse scenarios. Traditional handcrafted methods for palmprint recognition often fall short in representation capability, as they heavily depend on researchers' prior knowledge. Deep learning (DL) has been introduced to address this limitation, leveraging its remarkable successes across various domains. While existing surveys focus narrowly on specific tasks within palmprint recognition—often grounded in traditional methodologies—there remains a significant gap in comprehensive research exploring DL-based approaches across all facets of palmprint recognition. This paper bridges that gap by thoroughly reviewing recent advancements in DL-powered palmprint recognition. The paper systematically examines progress across key tasks, including region-of-interest segmentation, feature extraction, and security/privacy-oriented challenges. Beyond highlighting these advancements, the paper identifies current challenges and uncovers promising opportunities for future research. By consolidating state-of-the-art progress, this review serves as a valuable resource for researchers, enabling them to stay abreast of cutting-edge technologies and drive innovation in palmprint recognition.

\end{abstract}


%
\IEEEpeerreviewmaketitle

\section{Introduction}

\IEEEPARstart {B}{iometric} recognition technology has emerged as a prominent identity management method in recent years, with applications spanning diverse domains ~\cite{jain2021biometrics}. Its effectiveness relies on the uniqueness of physiological traits such as face~\cite{boutros2023synthetic}, iris~\cite{nguyen2024deep}, and 
palmprint~\cite{trabelsi2022efficient}, as well as behavioral traits like keystroke dynamics~\cite{alfalahi2022diagnostic}, gait~\cite{sepas2022deep}, and signature~\cite{li2024deep}. As a relatively new physiological modality, palmprint encompasses a rich array of distinctive features, including wrinkles, principal lines, intricate ridges, and fine-grained characteristics~\cite{rodriguez2019survey}. Palmprint is generally perceived as less intrusive than other biometric modalities, and the recognition process is notably user-friendly. These attributes make palmprint recognition a highly promising approach for achieving high accuracy and reliability in personal verification and identification applications. 

\begin{figure}
  \begin{center}
  \includegraphics[width=.6\columnwidth]{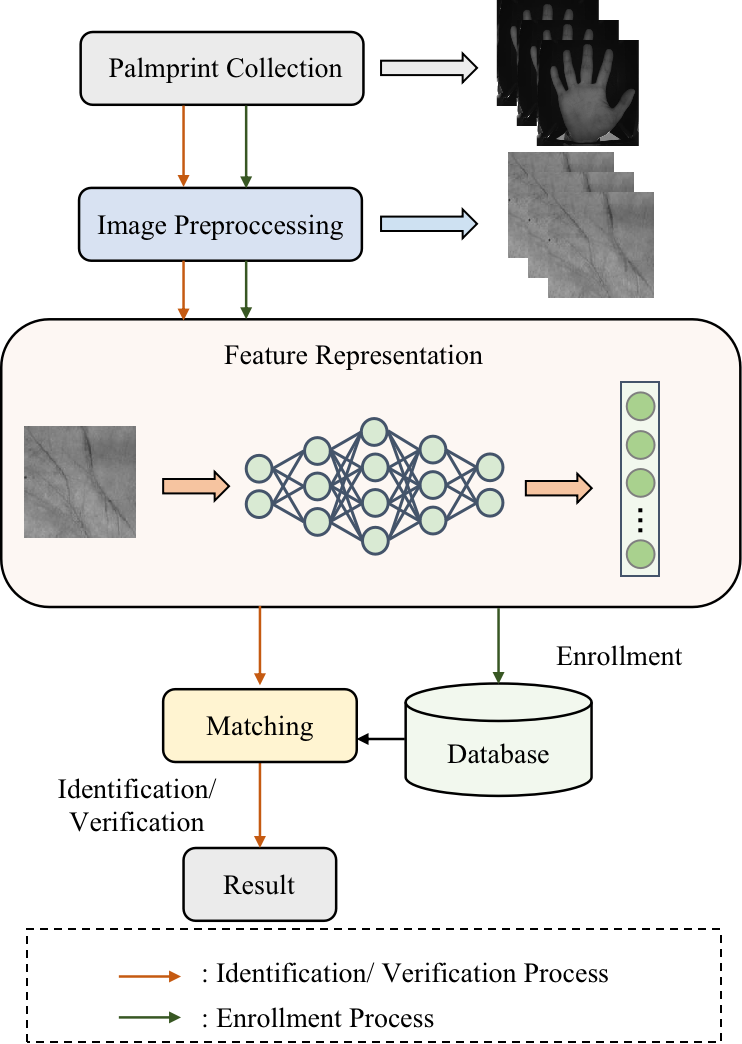}
  \vspace{-5pt}
  \caption{The pipeline of the DL-based palmprint recognition system.}
  \vspace{-25pt}
  \label{process}
  \end{center}
\end{figure}

\begin{table*}[!t]
\caption{Comparison between recent surveys and this paper}
\centering
\resizebox{.75\textwidth}{!}{
\begin{tabular}{p{6cm}llp{3cm}}
\hline
\textbf{Paper Title}   & \textbf{Year} & \textbf{Focus/Keywords}    & \textbf{Technique}                                                  \\ \hline
A survey of palmprint recognition~\cite{kong2009survey}    & 2009 & \begin{tabular}[c]{@{}l@{}}1. Feature extraction \\ 2. Acquistion devices\\ 3. Preprocessing\\ 4. Privacy protection\end{tabular}   & Traditional methods                                         \\ \hline
A comparative study of palmprint recognition algorithms~\cite{zhang2012comparative}    & 2012 & \begin{tabular}[c]{@{}l@{}}1. Feature extraction \\ 2. Low-resolution palmprint\end{tabular}    & Traditional methods  \\ \hline
A survey on minutiae-based palmprint feature representations, and a full analysis of palmprint feature representation role in latent identification performance~\cite{rodriguez2019survey} & 2019 & \begin{tabular}[c]{@{}l@{}}1. Latent palmprint identification\\ 2. Minutiae-based features\\ 4. Latent palmprint matching\end{tabular}      & Traditional methods     \\ \hline
Decade progress of palmprint recognition: A brief survey~\cite{zhong2019decade}                                                                                                               & 2019 & \begin{tabular}[c]{@{}l@{}}1. Preprocessing \\ 2. Feature extraction\end{tabular}  & Traditional methods                                         \\ \hline
\begin{tabular}[c]{@{}l@{}}Feature extraction methods for palmprint\\ recognition: A Survey and evaluation~\cite{fei2018feature}\end{tabular}                                                 & 2019 & \begin{tabular}[c]{@{}l@{}}1. Feature extraction \\ 2. Image analysis\end{tabular}   & Traditional methods     \\ \hline
\begin{tabular}[c]{@{}l@{}}Feature extraction for 3-D palmprint\\ recognition: A survey~\cite{fei2020feature}\end{tabular}   & 2020 & \begin{tabular}[c]{@{}l@{}} 1. 3-D palmprint recognition\\ 2. Feature extraction\\ 3. Data acquisition\\ 3. Preprocessing\end{tabular}   & Traditional methods           \\ \hline
\begin{tabular}[c]{@{}l@{}}Toward unconstrained palmprint recognition\\ on consumer devices: A literature Review~\cite{ungureanu2020toward}\end{tabular}      & 2020 & \begin{tabular}[c]{@{}l@{}} 1. Acquistion devices\\ 2. ROI extraction\\ 3. Feature extraction\end{tabular}        & Traditional methods and deep learning methods       \\ \hline
Multiview-learning-based generic palmprint recognition: A literature review~\cite{zhao2023multiview}      & 2023 & \begin{tabular}[c]{@{}l@{}}1. Multiview learning\\ 2. Palmprint recognition\\ 3. Open-set environments\end{tabular}    & Traditional methods and deep learning methods               \\ \hline

Deep learning in palmprint recognition: A comprehensive survey~(Ours)   & \_    & \begin{tabular}[c]{@{}l@{}}1. Preprocessing\\ 2. Feature extraction\\ 3. Closed-set and open-set recognition\\  4. Security and Privacy\\ 5. Other palmprint recognition tasks\end{tabular} & Mainly deep learning methods and several traditional methods\\ \hline
\end{tabular}
}
\vspace{-10pt}
\label{tab:review}
\end{table*}

The palmprint recognition process follows a structured pipeline comprising four essential steps: image acquisition, preprocessing, feature extraction, and matching, as depicted in Fig.~\ref{process}. Palmprint images are captured during enrollment, and regions of interest (ROI) are identified, making preprocessing a critical step to facilitate accurate feature extraction and matching. The extracted discriminative features are stored as templates during enrollment. The query features, derived from the same processing steps, are compared against stored templates to verify or determine an individual's identity. These foundational steps have shaped the evolution of palmprint recognition, advancing from statistical methods to deep learning-based approaches.

Palmprint recognition has evolved significantly across different technological eras. In its early stages, the field relied heavily on statistic-based approaches, which depended on manually designed methods to extract statistical information. However, these approaches often fell short of capturing the intricate texture details of palmprints. A major shift occurred in 2003, with researchers focusing on texture features extracted from palmprint images. Among the seminal works of this period was PalmCode~\cite{zhang2003online}. Introducing Competitive Code~(CompCode)~\cite{kong2004competitive} further advanced the field, highlighting the robustness of ordering features in palmprint recognition. This discovery spurred extensive research into developing powerful competition mechanisms to enhance recognition accuracy.

The advent of deep learning (DL) further transformed the field. While initial attempts with generic DL models produced suboptimal results, combining DL with traditional encoding techniques led to significant breakthroughs. These innovations addressed challenges unique to palmprint recognition, propelling the field toward robust and accurate solutions.

Several survey papers on palmprint biometrics have been published, primarily emphasizing traditional feature extraction techniques or specialized tasks. For instance, Zhong \textit{et al}.~\cite{zhong2019decade} provided an overview of palmprint recognition, addressing data collection, datasets, preprocessing, traditional feature extraction, matching strategies, and fusion techniques. Similarly, Fei \textit{et al}.~\cite{fei2018feature} categorized traditional feature extraction methods into four types, examining the theoretical foundations of extraction and matching approaches, particularly for 3D palmprint images. Ungureanu \textit{et al}.~\cite{ungureanu2020toward} investigated unconstrained palmprint recognition, focusing on ROI extraction methods, feature extraction strategies, and matching algorithms. Lately, Zhao \textit{et al}.~\cite{zhao2023multiview} highlighted recognition methods based on multi-view learning, discussing how integrating complementary features from multiple perspectives enhances recognition performance. 

However, unlike the more specialized reviews, this paper comprehensively explores diverse tasks, highlighting the transformative impact of DL technologies. It delves into emerging advancements in palmprint biometrics, including secure and privacy-preserving recognition, open-set recognition, cross-domain and cross-modality techniques, lightweight systems, and applications extending beyond identity recognition. A comparison with previous surveys is summarized in Table~\ref{tab:review}, highlighting the expansive scope of this review.

The main contributions of this paper can be summarized as follows:
\begin{itemize}
    \item This paper spotlights the revolutionary role of DL technologies, unveiling their transformative potential across a wide spectrum of tasks in palmprint biometrics.

    \item This paper explores emerging tasks in palmprint recognition that were not addressed in previous surveys.

    \item This paper summarizes the current research landscape and discusses potential future research directions in palmprint recognition.
    
\end{itemize}

\begin{figure}
  \begin{center}
  \includegraphics[width=\columnwidth]{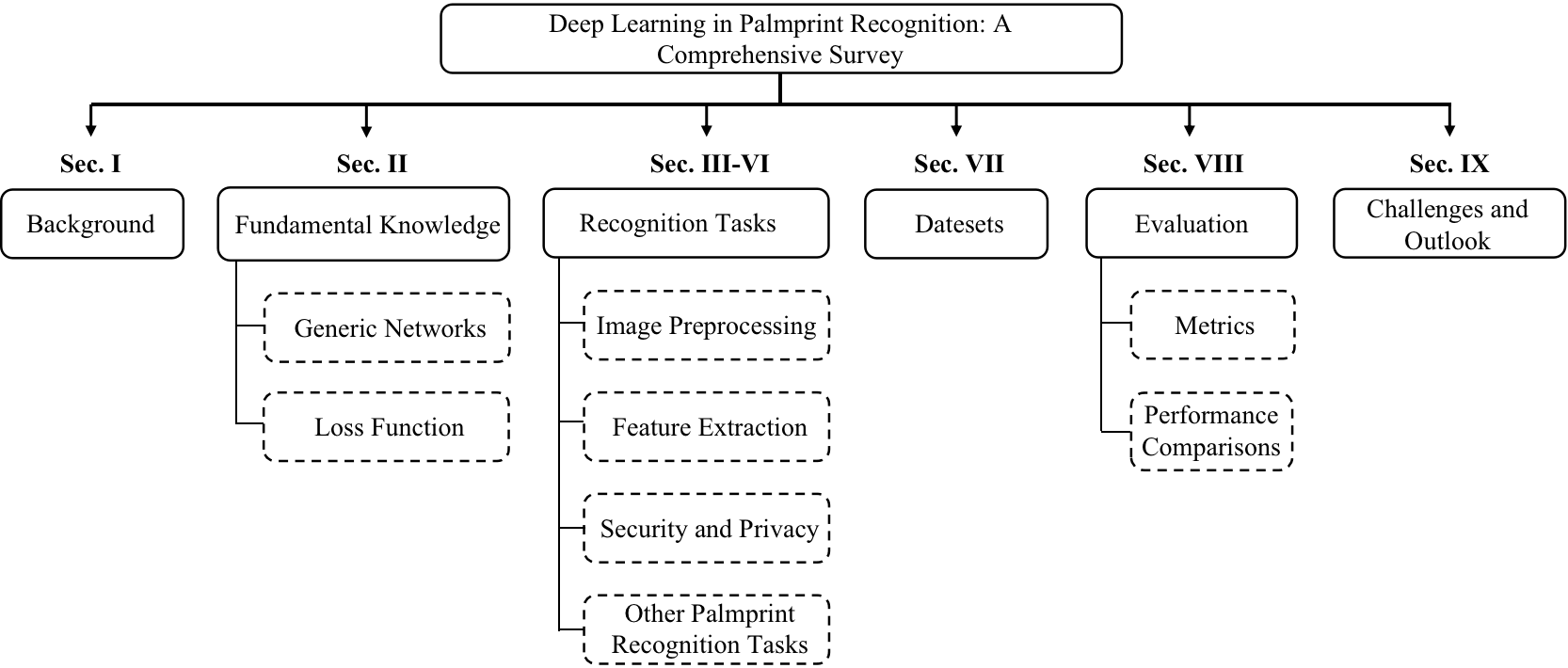}
  \vspace{-15pt}
  \caption{Overall structure of this paper.}
  \vspace{-20pt}
  \label{fig:organization}
  \end{center}
\end{figure}

As depicted in Fig.~\ref{fig:organization}, the structure of this paper is as follows: 
Section II overviews DL approaches in palmprint recognition, laying the groundwork for subsequent discussions. Section III highlights preprocessing techniques to enhance recognition accuracy. Section IV focuses on state-of-the-art feature extraction methods and advancements. Section V addresses security and privacy concerns. Section VI discusses specialized tasks in palmprint recognition with insights and comparisons. Section VII reviews commonly used datasets. Section VIII evaluates performance metrics and synthesizes experimental findings. Finally, Section IX summarizes contributions and suggests future research directions.

\section{Background Knowledge for DL in Palmprint Recognition}
Since 2006, DL has made breakthrough advances in artificial intelligence (AI), driving the prosperity of AI research and industry~\cite{gu2018recent}. In recent years, there has been a growing trend in conducting palmprint recognition research using DL, with early attempts demonstrating its potential~\cite{fei2022toward}. This section will introduce several generic neural networks, and loss functions widely used in the following tasks.

\subsection{Generic Neural Networks}

The adoption of DL, initially developed for general computer vision tasks, marked a significant shift in palmprint recognition beginning around 2015. These methods leveraged generic neural networks such as simple Convolutional Neural Networks (CNNs)~\cite{lecun1989handwritten}, Deep Belief Networks (DBNs)~\cite{hinton2006fast}, AlexNet~\cite{krizhevsky2017imagenet}, CNN-Fast~\cite{chatfield2014return}, VGG-16~\cite{simonyan2014very}, PCANet~\cite{chan2015pcanet}, Inception\_ResNet\_v1~\cite{szegedy2017inception}, Siamese Networks~\cite{chen2021exploring}, MobileNet-V2~\cite{sandler2018mobilenetv2}, ResNet~\cite{he2016deep}, Vision Transformers (ViT)~\cite{dosovitskiy2020image}, and others~\cite{jia20212d,jia2021performance}. These networks were adapted to address the unique challenges of palmprint recognition, such as capturing fine-grained textural features and handling variations in acquisition conditions.

Simple CNNs, characterized by their shallow depth and straightforward architecture, were among the first to be applied in this domain~\cite{jalali2015deformation}. Their simplicity facilitated quick training and low computational requirements, making them ideal for early experiments. Building on this, Xin \textit{et al.}~\cite{xin2015palmprint} explored using DBNs, generative models proficient at learning hierarchical feature representations, to extract meaningful patterns from palmprint images.

AlexNet, a pioneering DL model in large-scale image recognition, marked a turning point with its adaptation to palmprint recognition by Dian \textit{et al.}~\cite{dian2016contactless}. AlexNet’s robust feature extraction capabilities enabled significant improvements in accuracy, revealing the potential of DL in capturing the intricate patterns necessary for reliable biometric systems. Similarly, VGG-16~\cite{simonyan2014very} gained traction in palmprint recognition for effectively learning hierarchical features, contributing to superior recognition performance~\cite{zhao2020deep, matkowski2019palmprint}.

PCANet, a hybrid method incorporating Principal Component Analysis (PCA) for unsupervised feature extraction, provided a bridge between traditional handcrafted features and deep learning. PCANet efficiently captured discriminative features from palmprint images by combining PCA with convolution operations, offering an accessible yet powerful alternative~\cite{meraoumia2017improving}.

Advanced architectures like Inception\_ResNet\_v1 and ResNet brought transformative improvements to palmprint recognition. Inception\_ResNet\_v1 leveraged multi-scale feature extraction and residual connections to achieve high accuracy~\cite{zhang2018palmprint}. ResNet, on the other hand, addressed the vanishing gradient problem in deep networks, allowing researchers to train much deeper models without overfitting, yielding impressive results~\cite{fei2018feature}.

Siamese Networks also emerged as a valuable tool in biometric verification, including palmprint verification~\cite{zhong2018palmprint}. These networks, consisting of two identical subnetworks, are designed to measure similarity between input pairs, making them particularly effective for verification tasks.

Innovative adaptations have continued to push the boundaries of palmprint recognition. Xu \textit{et al.}~\cite{xu2021effective} proposed a contactless palmprint recognition algorithm utilizing a Spatial Transformer Network (STN) for precise alignment, followed by CNN-based learning and classification.

More recently, Grosz \textit{et al.}~\cite{grosz2024mobile} integrated ViT with CNNs to combine complementary local and global feature extraction, creating Palm-ID, an end-to-end mobile-based palmprint recognition system. This hybrid approach highlights the growing trend of leveraging transformer-based architectures for biometric tasks.

The progressive adaptation of these generic neural networks to palmprint recognition reflects the versatility of DL and its capacity to evolve in response to domain-specific challenges. Researchers have significantly advanced the field by combining foundational models with innovative strategies.

\subsection{Loss Function}
In DL, loss functions are crucial in shaping model optimization, boosting recognition accuracy, and reinforcing robustness. Tailored to meet specific task requirements, these functions are extensively used across different stages, including feature learning, matching, and classification. Below, we highlight some of the most widely employed loss functions in optimizing DL models for palmprint recognition.

(1) Cross-Entropy Loss~\cite{goodfellow2016deep}:
Cross-entropy loss is the most commonly used loss function for classification tasks, especially when dealing with multi-class classification problems in DL models. Cross-entropy loss quantifies the discrepancy between the predicted and true class probability distributions.

\begin{equation}
\mathcal{L}_{\text{CE}} = - \sum_{i} y_i \log(\hat{y}_i),
\end{equation}
where \( y_i \) is the true class distribution (class label) and \( \hat{y}_i \) is the predicted probability distribution from the model.

(2) Contrastive Loss~\cite{chopra2005learning}:
Contrastive loss is a widely used loss function in metric learning, particularly in Siamese networks. Its objective is to minimize the distance between similar samples while maximizing the distance between dissimilar ones. By leveraging contrastive loss, networks can learn robust feature representations, ultimately enhancing recognition accuracy.

\begin{equation}
\mathcal{L}_{\text{contrastive}} = \frac{1}{2} \left( t \cdot D^2 + (1 - t) \cdot \max(0, m - D)^2 \right),
\end{equation}
where \( t \) is 1 for similar pairs and 0 for dissimilar pairs, \( D \) is the Euclidean distance between the embeddings, and \( m \) is the margin.

(3) Triplet Loss~\cite{schroff2015facenet}:
Triplet loss is also for metric learning. It trains the network to understand the relationships among three key samples: an anchor, a positive (from the same class as the anchor), and a negative (from a different class). The goal is to ensure that similar samples are closer in feature space than dissimilar ones.

\begin{equation}
\mathcal{L}_{\text{triplet}} = \max \left( D(\mathbf{a}, \mathbf{p}) - D(\mathbf{a}, \mathbf{n}) + \alpha, 0 \right),
\end{equation}
where \( D(\mathbf{a}, \mathbf{p}) \) is the distance between anchor \( \mathbf{a} \) and positive sample \( \mathbf{p} \), \( D(\mathbf{a}, \mathbf{n}) \) is the distance between anchor \( \mathbf{a} \) and negative sample \( \mathbf{n} \), and \( \alpha \) is the margin.



(4) Focal Loss~\cite{ross2017focal}:
Focal loss is a specialized function crafted to tackle class imbalance challenges, particularly in scenarios with a pronounced disparity among identity samples. It prioritizes difficult-to-classify samples by incorporating a modulating factor, ensuring the model pays greater attention to these challenging instances. In applications such as contactless palmprint recognition~\cite{zhang2024hgaiqa}, focal loss proves invaluable, bolstering the model's robustness and overall performance.

\begin{equation}
\mathcal{L}_{\text{focal}} = - \beta (1 - \hat{p})^\gamma \log(\hat{p}),
\end{equation}
where \( \hat{p} \) is the predicted probability for the class, \( \beta \) is a scaling factor, and \( \gamma \) is the focusing parameter.

(5) Margin Loss~\cite{mikolajczyk2005performance}:
Margin loss is widely used in identity verification tasks to create a discriminative feature embedding space. Reducing the distance between samples of the same identity and increasing the separation between samples of different identities enhances the network's ability to distinguish between biometric data. This loss function is frequently applied in tasks such as palmprint verification~\cite{zhong2019centralized}.

\begin{equation}
\mathcal{L}_{\text{margin}} = \max(0, D(\mathbf{a}, \mathbf{p}) - D(\mathbf{a}, \mathbf{n}) + \alpha),
\end{equation}
where \( D(\mathbf{a}, \mathbf{p}) \) is the distance between anchor \( \mathbf{a} \) and positive sample \( \mathbf{p} \), \( D(\mathbf{a}, \mathbf{n}) \) is the distance between anchor \( \mathbf{a} \) and negative sample \( \mathbf{n} \), and \( \alpha \) is the margin.

(6) Mean Squared Error (MSE) Loss~\cite{lecun2015deep}:
MSE is a widely used loss function in regression and generation tasks. In palmprint recognition methods that utilize Generative Adversarial Networks (GANs) or autoencoders, the MSE loss is often employed to enhance feature reconstruction and improve generated output quality~\cite{liu2021palmprint}.

\begin{equation}
\mathcal{L}_{\text{MSE}} = \frac{1}{N} \sum_{i=1}^{N} (z_i - \hat{z}_i)^2,
\end{equation}
where \( z_i \) is the true value and \( \hat{z}_i \) is the predicted value.

\section{Image Preprocessing}
The preprocessing stage in palmprint recognition is crucial. It focuses on extracting the Region of Interest (ROI) and improving the image quality. These steps are essential, as they directly influence the accuracy and reliability of the subsequent feature extraction process.

\subsection{Region of Interest Extraction}
ROI extraction methods are broadly classified into two categories: standard extraction techniques~\cite{dalal2005histograms} and advanced DL-based approaches~\cite{liu2020contactless}.

One commonly used ROI extraction technique is the Tangent-Based method~\cite{nalamothu2021review}, which identifies valley points between fingers by constructing tangent lines across the palmprint gaps. These points establish a reliable coordinate system. Zhang \textit{et al}.~\cite{zhang2017towards} extended this method to contactless scenarios using a specialized acquisition device.

Another standard approach is the Contour Profile Distance Distribution-Based method~\cite{lin2005palmprint}. It uses a scanner to capture the hand image, identifies hand boundaries, and determines a reference point at the wrist's intersection. Euclidean distances from the reference to boundary pixels form a contour profile curve, with local minima marking the ROI sub-image.

The methods above perform well under controlled environmental conditions but are highly sensitive to pose variations. DL-based approaches have recently emerged to address these limitations. Bao \textit{et al}.~\cite{bao2016extracting} introduced a two-stage shallow neural network system: the first network classifies the palm image as either a left or right hand, while the second network detects the coordinates of three valley points. A similar approach was proposed by Izadpanahkakhk \textit{et al}.~\cite{izadpanahkakhk2018deep}, which outputs the coordinates of a corner point along with the width and height of the ROI area. However, both methods were evaluated only on constrained datasets. 

Advancing the field further, Li \textit{et al}.~\cite{li2021bpfnet} proposed the Bimodal Palmprint Fusion Network (BPFNet), which performs ROI localization, alignment, and bimodal image fusion in an end-to-end manner. BPFNet employs a detection network to directly regress the rotated bounding box based on the point of view while predicting image disparity.

Luo \textit{et al}.~\cite{luo2021robust} developed a model comprising a palm detection module and a key point detection module. After identifying palm key points, the system establishes a coordinate framework to extract the ROI. An auxiliary network estimates the palm's angle to enhance model convergence and accuracy. Additionally, Liang \textit{et al}.~\cite{liang2023pklnet} introduced the Palm Keypoint Localization Network (PKLNet), which integrates information from the hand region, palm boundary, and finger valley edges, achieving robust and precise keypoint localization for effective ROI extraction. 

Lin \textit{et al}.~\cite{lin2024unconstrained} presented a lightweight network-based method for palmprint ROI extraction. It first used YOLOv5-lite for palm detection and initial localization, removing background interference. Then, an improved UNet model is employed for keypoint detection, reducing parameters and boosting performance compared to the original UNet. 

Most recently, Chai \textit{et al}.~\cite{chai2024joint} proposed an automated and flexible ROI extraction method for complex scenarios based on Finger Valley Points-Free adaptive ROI detection. This method combines cross-dataset hand shape semantic transfer and the constrained palm inscribed circle search, enabling excellent hand segmentation and precise ROI extraction.

\subsection{Palmprint Image Quality}
The quality of palmprint images is critical in ensuring accurate recognition, spurring the development of various techniques to enhance low-quality biometric images. Research consistently highlights the significant impact of image quality on recognition performance, leading to exploring methods like denoising and image super-resolution (SR) to address these challenges effectively.

Denoising has emerged as a key focus in palmprint recognition research. For instance, Arora \textit{et al}.~\cite{arora2023evaluation} examined the effects of noise on biometric image quality and proposed a DL-based method leveraging CNN architectures such as InceptionV3, VGG16, and ResNet50 to classify denoised images. The proposed approach demonstrated its efficacy through rigorous performance evaluations on widely recognized benchmarks.

While palmprint recognition systems thrive on high-quality images, real-world scenarios often involve capturing low-quality samples. To address this, image SR techniques have proven effective in enhancing image quality. Wang \textit{et al}.~\cite{wang2024dense} introduced the Dense Hybrid Attention (DHA) network, tailored specifically for palmprint image SR. This innovative method begins by generating high-dimensional shallow representations via a single convolutional layer. Subsequently, it employs parallel CNN and Transformer-based branches to jointly learn local and global palmprint features, restoring distinctive palmprint-specific attributes.

Another pressing challenge is assessing image quality, especially in practical situations with limited labeled quality data. Zou \textit{et al}.~\cite{zou2023unsupervised} tackled this issue with their Pseudo-Label Generation and Ranking Guidance (PGRG) framework for unlabeled palmprint image quality assessment (PIQA). This two-stage method estimates the recognizability of palmprint images to generate pseudo-labels. It then produces pseudo-images and assigns quality rankings for pre-training, employing a label-based ranking loss to help the model discern relative quality relationships among pseudo-labels.

These advancements highlight the importance of improving and evaluating image quality in palmprint recognition systems. By tackling challenges such as noise reduction, resolution enhancement, and quality assessment, researchers continue to strengthen the robustness and reliability of these systems.

\begin{figure*}
  \begin{center}
  \includegraphics[width=.75\textwidth]{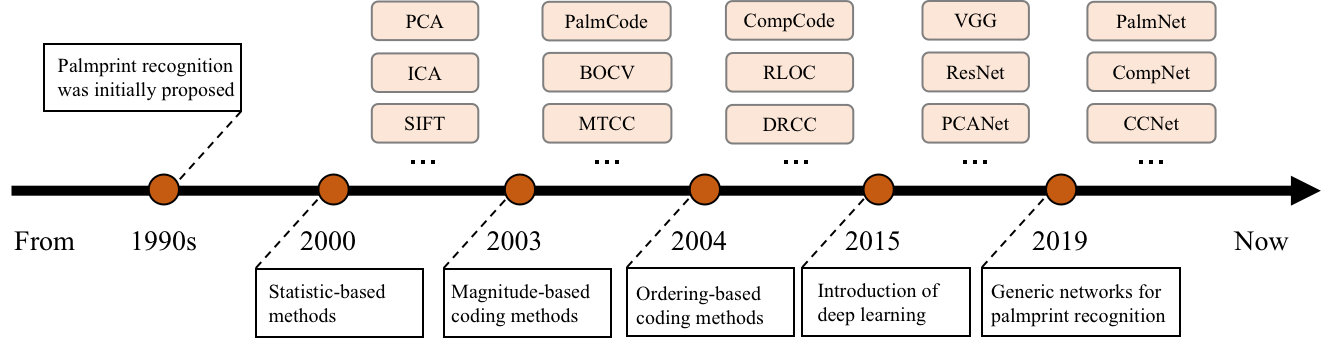}\\
  \caption{The evolution of palmprint feature extraction methodologies.}
  \label{history}
  \vspace{-20pt}
  \end{center}
\end{figure*}

\section{Feature Extraction}

Since the inception of palmprint recognition, researchers have explored a wide range of approaches, including statistical methods, coding-based techniques, and, more recently, DL-based strategies. The evolution of these methodologies is illustrated in Fig.~\ref{history}, showcasing key advancements in the field. 

Statistical-based methods employ statistical transformation techniques to extract meaningful features from palmprint images. These methods operate directly on the raw palmprint image or leverage transformed representations, such as coefficients obtained from Fourier or wavelet transforms, to derive distinctive feature sets. Prominent approaches include Principal Component Analysis (PCA), which reduces dimensionality by identifying the directions of maximum variance in the data, effectively capturing the most critical patterns~\cite{bai20173d}. 

On the other hand, Independent Component Analysis (ICA) focuses on separating statistically independent components, enabling robust feature extraction even in the presence of noise~\cite{malviya2011verification}. Additionally, the Scale-Invariant Feature Transform~(SIFT) has been widely adopted for its ability to detect and describe local features invariant to scale, rotation, and minor image distortions~\cite{wu2014sift}. 

Encoding-based methods in palmprint recognition can be classified into magnitude and ordering feature-based methods. Magnitude feature-based methods apply predefined coding rules directly to the extracted features, creating compact and discriminative templates. Notable examples include Palmcode~\cite{zhang2003online}, which leverages Gabor filters to encode phase information, BOCV~\cite{guo2009palmprint}, which utilizes binary orientation co-occurrence vectors, and MTCC~\cite{yang2023multi}, a recent technique focusing on multi-task coding for enhanced accuracy.

In contrast, ordering feature-based methods encode information by comparing magnitude responses along different orientations. Key methods in this category include CompCode~\cite{kong2004competitive}, which uses competitive coding to identify dominant orientations; RLOC~\cite {jia2008palmprint}, which emphasizes robust line orientation coding; DOC~\cite{fei2016double}, which combines dual orientation coding for greater precision, and DRCC~\cite{xu2016discriminative}, which incorporates discriminative ridge coding for improved distinctiveness.

While these methods have demonstrated commendable performance, their reliance on predefined coding rules and the researchers' domain knowledge imposes inherent limitations, resulting in moderate robustness under challenging or less-controlled scenarios.

The advent of DL has revolutionized palmprint recognition, leading to the development of numerous methods based on deep neural networks that deliver impressive performance. Unlike traditional approaches, DL-based methods can establish end-to-end frameworks, offering enhanced robustness and adaptability. 

While early generic DL networks were applied to palmprint recognition, their performance often fell short due to the unique challenges presented by palmprint data.

Generic deep learning models have surpassed traditional handcrafted methods in certain areas, they are not optimized for the distinctive traits of palmprint images. These challenges include significant intra-class variations, such as differing hand positions, lighting conditions, skin textures, and pronounced inter-class similarities, where palmprints from different individuals may appear deceptively alike. Generic architectures often fail to capture the unique textures and features of palmprints, emphasizing the need for advanced, specialized DL approaches to enhance recognition accuracy and efficiency.

Researchers have introduced specialized neural network architectures tailored specifically for palmprint recognition to overcome these obstacles. These custom designs address the intricacies of palmprint images, improving feature extraction and enabling more accurate and reliable recognition, even in complex and unconstrained scenarios. This tailored approach has propelled DL-based methods to the forefront of palmprint recognition research.

Unlike traditional palmprint feature extraction methods, which primarily emphasize enrollment followed by testing~(verification or identification), DL-based techniques incorporate an additional phase: model training. This inclusion creates two distinct scenarios:

\begin{figure}[!t]
  \begin{center}
  \includegraphics[width=.9\columnwidth]{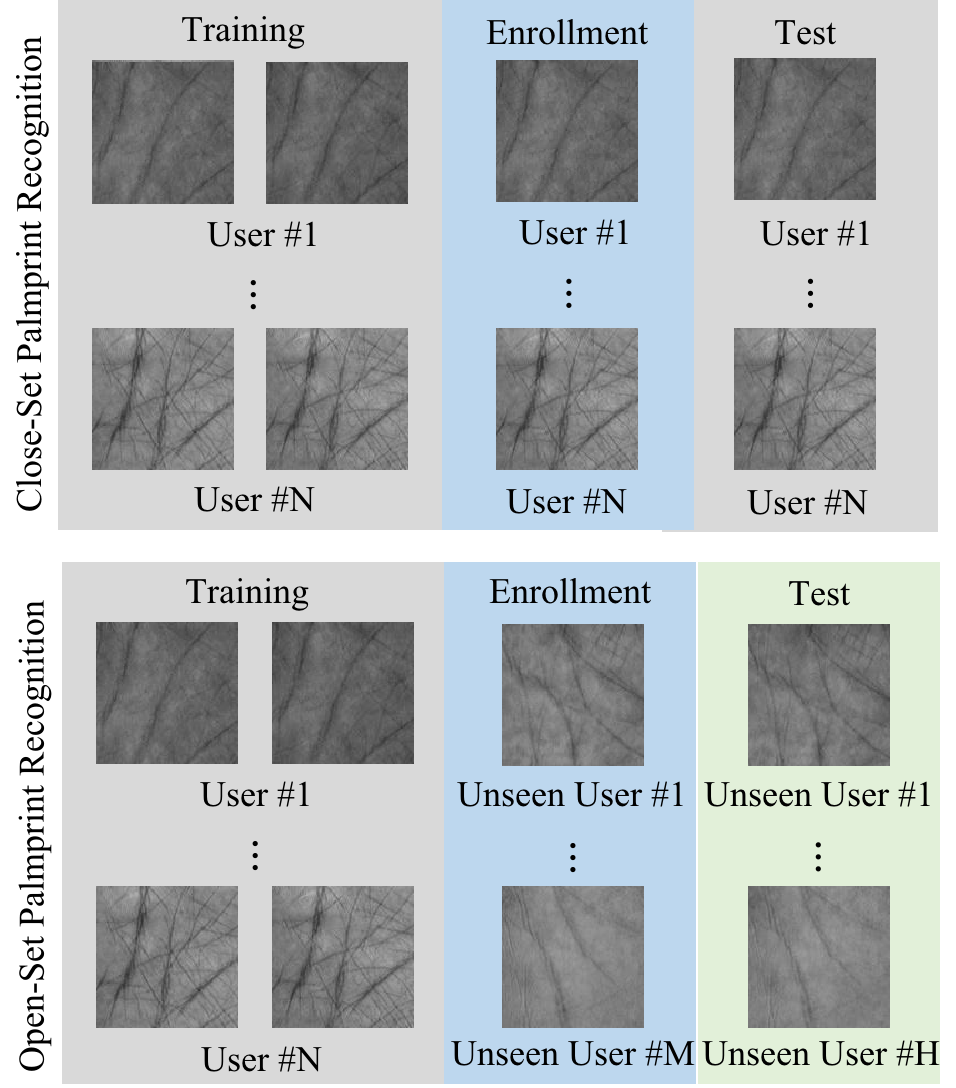}
  \caption{The illustration of closed-set and open-set palmprint recognition scenario.}
  \vspace{-20pt}
  \label{close-open}
  \end{center}
\end{figure}

\begin{itemize}

\item \textbf{Closed-Set Palmprint Recognition:} In closed-set palmprint recognition, both the enrolled and testing identities are fully predefined and included in the training set. This approach focuses on achieving highly accurate recognition within a controlled group of known individuals. Since all identities are accounted for during training, the task becomes one of matching palmprints to a specific identity within the predetermined group.

\item \textbf{Open-Set Palmprint Recognition:} Open-set palmprint recognition presents a more intricate and practical challenge than its closed-set counterpart. The enrolled and testing identities are entirely separate from the training set in this scenario. As a result, the system must process palmprints from individuals it has never seen before, making the task significantly more challenging. This setting mirrors real-world scenarios, where recognition systems must adapt to dynamic environments and handle unseen users seamlessly. In the palmprint research community, this open-set paradigm is often called cross-dataset recognition~\cite{shao2024learning} or cross-id recognition~\cite{yang2023comprehensive}. The distinctions between the two scenarios are illustrated in Fig.~\ref{close-open}.
\end{itemize}

\subsection{Closed-Set Palmprint Recognition}

One of the early contributions under this category came from 
Kumar \textit{et al}.~\cite{kumar2016identifying} took a novel approach by exploring cross-hand palmprint matching between right and left images. Employing CNNs, they achieved commendable results. Samai \textit{et al}.~\cite{samai20182d} proposed a dual-dimensional model for 2D and 3D palmprint recognition. By converting 3D palmprint data into grayscale images using Mean Curvature (MC) and Gauss Curvature (GC), they extracted features using a Discrete Cosine Transform Network (DCT Net). This innovative method leveraged matching score-level fusion for efficient biometric identification.

A significant breakthrough came with the work of Zhong \textit{et al}.~\cite{zhong2018palmprint}, who introduced a palmprint recognition algorithm based on the Siamese network. Utilizing two VGG-16 networks with shared parameters, their model extracted features from input palmprints and determined their similarity based on convolutional features.
Izadpanahkakhk \textit{et al}.~\cite{izadpanahkakhk2018deep} employed transfer learning for palmprint verification, extracting ROI and features jointly. They combined a pre-trained CNN with an SVM classifier, delivering strong ROI segmentation and recognition performance.

In another direction, Xie \textit{et al}.~\cite{xie2018palmprint} applied CNNs for gender classification using palmprint images. By fine-tuning a pre-trained VGGNet on palmprint datasets, they demonstrated excellent results in this novel application of palmprint data.
Shao and Zhong~\cite{shao2019few} presented a few-shot palmprint recognition model using graph neural networks (GNNs). In their framework, CNN-extracted palmprint features were treated as nodes in the GNN, with edges representing similarities between nodes, enabling efficient recognition with limited training samples.

Further innovation came from Shao \textit{et al}.~\cite{shao2019efficient}, who proposed a deep recognition approach integrating hash coding and knowledge distillation. By extending Deep Hash Networks~(DHN) to binary code generation, their method optimized storage and accelerated matching processes. Zhao \textit{et al}.~\cite{zhao2019joint} proposed a joint deep convolutional feature representation in hyperspectral palmprint recognition. Their CNN stack effectively processed 53 spectral bands, achieving promising performance. 

Genovese \textit{et al}.~\cite{genovese2019palmnet} introduced PalmNet, a deep recognition algorithm incorporating Gabor responses and PCA. This multi-stream CNN captured diverse features through attention mechanisms, including textures and unique details. Hussein \textit{et al}.~\cite{hussein2020multimodal} developed a texture-based recognition approach using a statistical gray-level co-occurrence matrix (GLCM). This was paired with a probabilistic neural network (PNN) to measure system performance, enhancing identification accuracy. Expanding on periodic feature sampling, Zhao \textit{et al}.~\cite{zhao2020joint} utilized CNNs for extracting discriminative convolutional features across palmprint images' local regions.

In recent advancements, Liang \textit{et al}.~\cite{liang2021compnet} proposed a Competitive Convolutional Neural Network (CompNet), a contactless recognition model resilient to illumination and scale variations. CompNet efficiently captured rich directional information utilizing multi-scale competitive blocks, proving highly effective on small-scale datasets. Most recently, Yang \textit{et al}.~\cite{yang2023co} introduced CO3Net, which employed multiscale learnable Gabor filters to enhance feature representation and discrimination. Additionally, Yang \textit{et al}.~\cite{yang2023comprehensive} developed CCNet, integrating spatial and channel competition mechanisms for efficient end-to-end palmprint information capture, setting a new benchmark in palmprint recognition.

\subsection{Open-Set Palmprint Recognition}

Palmprint recognition algorithms were traditionally tailored for closed-set scenarios. Adding new users demanded substantial time and effort to update the models. This challenge highlighted the need for methods that could effortlessly integrate new users without extensive retraining, leading to the development of open-set palmprint recognition.

Leveraging the shift toward open-set recognition, Zhong \textit{et al}.~\cite{zhong2019centralized} proposed Centralized Large Margin Cosine Loss (C-LMCL) to enhance class separation and improve open-set recognition. Shao \textit{et al}.~\cite{shao2020effective} introduced Deep Ensemble Hashing (DEH), combining weak feature extractors through online gradient boosting to boost recognition performance. 

Li \textit{et al}.~\cite{li2022row} introduced Row-wise Sparse Binary Feature Learning (RsBFL), which extracts binary features directly from image pixels, demonstrating strong generalization in open-set palmprint recognition. This approach was later extended to Row-Sparsity Binary Feature Learning for Multi-View Representation (RsBFL\_Mv)~\cite{li2024discriminative}, which harnesses intra-class and inter-class information across multiple views to enhance feature learning, providing a robust solution for open-set recognition.

Building on advancements in open-set palmprint recognition, Shao \textit{et al}.~\cite{shao2022towards} introduced the Weighted-based Meta Metric Learning (W2ML) framework, which uses meta-learning to enhance generalization and extract highly discriminative features through an end-to-end process. Complementing this, Shao \textit{et al}.~\cite{shao2021towards} proposed the Joint Pixel and Feature Alignment (JPFA) framework. This two-stage approach reduces pixel-level dataset gaps with deep style transfer and aligns feature distributions using domain adaptation.

Continuing this progression, Shao \textit{et al}.~\cite{shao2023multi} developed the Distilling from Multi-Teacher (DFMT) method. DFMT integrates knowledge distillation and domain adaptation by pairing source datasets with multiple targets and employing teacher feature extractors for adaptive knowledge capture. A student feature extractor then learns this knowledge through multilevel distillation losses, facilitating effective knowledge transfer across diverse datasets.

Shao \textit{et al}.~\cite{shao2024learning} recently introduced the Palmprint Data and Feature Generation (PDFG) method for Cross-Dataset Palmprint Recognition with Unseen Targets (CDPR-UT). This approach boosts model adaptability to unseen datasets by using a Fourier-based data augmentation technique to create diverse training data and employing feature-level losses to reduce shifts between source and augmented datasets, ensuring adaptive feature extraction.


\section{Security and Privacy of Palmprint Recognition}

The practical deployment of biometric systems demands careful attention to their performance, user acceptability, and resistance to circumvention. The rapid advancement of digital connectivity further amplifies the importance of robust privacy protection, especially for sensitive biometric information. This section explores the privacy concerns associated with palmprints from two key perspectives: the challenges posed by potential attacks and the countermeasures.

\subsection{Attacks on Palmprint Recognition Systems}

\subsubsection{Palmprint Template Correlation Attacks}
Palmprint template correlation attacks exploit inherent statistical correlations within palmprint templates to compromise palmprint recognition systems. Researchers have developed various attack strategies to assess and enhance the security of palmprint recognition systems.

Leng~\textit{et al}.~\cite{leng2011dual} first revealed the statistical correlation inherent in coding-based palmprints and devised a cross-dataset attack algorithm to evaluate the security vulnerabilities of such systems.
Building upon this, Zhu~\textit{et al}.~\cite{zhu2023multi} introduced a joint attack and defense framework for multi-spectral palmprints. By leveraging information from diverse sources, their approach effectively reduced the correlation between projected features, disrupting the discriminative capabilities of palmprint models and limiting their representation and classification effectiveness.

Further exploring attack strategies, Yang~\textit{et al}.~\cite{yang2023two} proposed two style transfer methods to reconstruct palmprints. Their approach utilized the strong correlation between templates and their original texture information to enable online attacks.

\subsubsection{Palmprint Reconstruction Attacks}

Palmprint reconstruction attacks involve generating synthetic palmprint images from stored biometric templates to gain unauthorized access to recognition systems. By exploiting the information within these templates, attackers can reconstruct images that closely resemble the original palmprints, posing a significant security threat.

Yue \textit{et al}.~\cite{sun2022reinforced} introduce two palmprint reconstruction attack techniques: Modification Constraint within Neighborhood~(MCwN) and Batch Member Selection (BMS). These methods improve existing attacks by enhancing reconstructed palmprint images' naturalness, visual quality, and completeness. The techniques iteratively modify easily obtainable palmprint images using deep reinforcement learning to reduce matching distance while maintaining image quality.

Wang \textit{et al}.~\cite{wang2020palmprint} demonstrated a brute-force approach, using DCGAN to generate multiple palmprint images and verifying them sequentially until a match was found, showcasing the ability to quickly and effectively reconstruct high-quality images. Building on this, Yan \textit{et al}.~\cite{yan2024toward} introduced a more advanced black-box attack method, leveraging Progressive GAN (ProGAN) to generate highly realistic palmprint images from a single template.

\subsubsection{Spoofing attacks in Palmprint Recognition Systems}

Spoofing attacks, also known as presentation attacks, involve presenting counterfeit palmprint images to the system's sensor, aiming to deceive the biometric system into granting unauthorized access. These attacks exploit the system's reliance on authentic palmprint features, presenting fabricated or altered images that mimic genuine palmprints. 

Kanhangad \textit{et al}.~\cite{kanhangad2015anti} investigated spoof attacks on palmprint verification systems, focusing on display-based and print-based methods. Their research highlights the vulnerabilities of palmprint recognition systems to various spoofing techniques.

Sun and Wang~\cite{sun2022presentation} created six palmprint presentation attack datasets, employing physical display carriers such as photos and monitors, to assess the effectiveness of these attacks. Their experiments on texture coding-based and DL-based recognition methods revealed that spoofing attacks using stolen original palmprint images pose a significant threat due to their high success rates.

These diverse attacks underscore the evolving landscape of palmprint security challenges and highlight the need for robust defenses to mitigate these sophisticated attack strategies.

\subsection{Countermeasures}

In response to the growing threats to palmprint recognition systems, researchers have developed countermeasures to enhance palmprint recognition systems security. 

\subsubsection{Privacy-Preserving Palmprint Recognition Systems}

Privacy-preserving palmprint recognition systems ensure the secure handling of biometric data, shielding it from breaches or misuse during the recognition process. By leveraging advanced techniques such as homomorphic encryption, federated learning, watermarking, and synthetic palmprint generation, these systems safeguard privacy and strengthen the overall security of palmprint recognition technology. 

Guo \textit{et al}.~\cite{guo2023homomorphic} introduced a homomorphic encryption framework that encrypts palmprint recognition networks layer by layer, ensuring secure processing of images and network keys throughout the recognition process. Similarly, Tao \textit{et al}.~\cite{tao2020optical} developed an encryption system combining biometric keys, Singular Value Decomposition (SVD), and Fresnel transforms to protect palmprint images effectively.  

To address privacy concerns during training, federated learning has emerged as a decentralized solution that enables local training without sharing sensitive palmprint data. Shao \textit{et al}.~\cite{shao2023privacy} proposed Federated Metric Learning (FedML), which allows collaborative model training across multiple clients while ensuring data privacy and isolation. Complementing this, Yang \textit{et al}.~\cite{yang2024physics} introduced a spectral consistency loss function to align local and global model parameters, preventing model drift and maintaining global consistency.  

Liu \textit{et al}.~\cite{liu2022data} developed the Dynamic Random Invisible Watermark Embedding (DRIWE) model, which embeds watermarks into palmprint images’ regions of interest. This approach safeguards data during storage and transmission while ensuring accurate recognition by extracting the watermark before processing.  

On a different front, synthetic palmprint data offers a promising solution for privacy preservation by mimicking the statistical properties of real data, thereby eliminating the need for actual samples and reducing the risk of breaches or misuse. This innovative approach safeguards privacy and facilitates palmprint biometric system development, testing, and training. Shen \textit{et al}.~\cite{shen2023rpg} introduced RPG-Palm, a model synthesizing palmprints across multiple identities. It enhances intra-class diversity through a conditional modulation generator while maintaining identity consistency with identity-aware loss. 

Similarly, Jin \textit{et al}.~\cite{jin2024pce} proposed the Palm Crease Energy~(PCE) domain, which bridges Bézier curves and palmprints using a two-stage model. The first stage generates PCE images (creases) from Bézier curves, and the second stage produces palmprints (textures) from these PCE images.

These privacy-preserving innovations collectively enhance the security and reliability of palmprint recognition systems, addressing critical vulnerabilities across various stages of data processing.  

\subsubsection{Cancelable Palmprint Recognition System}

Cancelable palmprint technology, a biometric template protection~(BTP) method, adheres to the ISO/IEC 30136 standard~\cite{isoiec30136} by fulfilling key criteria: \textit{cancelability} for replacing compromised templates, \textit{unlinkability} for generating unique and non-correlatable templates, \textit{irreversibility} to protect user privacy, and high \textit{accuracy performance} to preserve recognition precision. 

Unlike traditional systems, cancelable approaches utilize user-specific tokens to apply non-invertible transformations to palmprint features, enabling secure template replacement and unlinkability. This approach effectively mitigates risks such as correlation and template reconstruction attacks, reinforcing the security and integrity of biometric systems.

An early instance is PalmHashing, introduced by Connie \textit{et al}.~\cite{connie2005palmhashing}, provided revocable palmprint templates, allowing users to replace compromised data. Building on this, Yang \textit{et al}.~\cite{yang2024dual} proposed a dual-level framework combining competition hashing for tokenized template generation with a negative dataset~(NDB) for enhanced protection, reducing data leakage by eliminating the need to store templates identical to those used for verification.  

Other approaches focus on efficiency and hybrid security. Siddhad \textit{et al}.~\cite{siddhad2020cancelable} utilized autoencoders to create low-dimensional templates, reducing storage size to 25\% of the original while maintaining performance. Sardar \textit{et al}.~\cite{sardar2022secure} introduced a hybrid scheme combining cancelable biometrics with bio-cryptography techniques, enhancing data security and privacy. Similarly, Khan \textit{et al}.~\cite{khan2024deep} proposed a method incorporating a deep attention network and randomized hashing, dynamically generating diverse templates through improved matrix multiplication using the LTTS random key and palmprint features.  

These diverse methods illustrate the evolution of cancelable palmprint technology, showcasing advancements in privacy, security, and efficiency for palmprint recognition systems.


\subsubsection{Anti-Spoofing and Liveliness Detection}

With the rise of spoofing techniques, palmprint verification systems require innovative solutions to maintain security and reliability. Researchers have explored diverse strategies, from leveraging surface properties to incorporating multispectral imaging, each contributing unique strengths to combat presentation attacks.  
  
Kanhangad \textit{et al}.~\cite{kanhangad2015anti} laid the groundwork by focusing on surface reflectance to counter display and print-based spoofing. By analyzing hand images through statistical features derived from pixel intensities and wavelet coefficients, their method introduced a foundational approach to detecting fake palmprints. 

Building on this, Aishwarya \textit{et al}.~\cite{aishwarya2016palm} enhanced the field by integrating biometric trait verification and Weber's local descriptor for feature refinement. This method complemented spoof detection with Euclidean distance-based authentication, though its liveness detection details remained unspecified, leaving room for further innovation. Expanding on vulnerability insights, Bhilare \textit{et al}.~\cite{bhilare2018study} delved deeper into presentation attacks. Their robust anti-spoofing solution advanced Kanhangad’s reflectance-based ideas, incorporating refined statistical analysis to accurately distinguish genuine from fake palmprints.  

Sugimoto \textit{et al}.~\cite{sugimoto2020liveness} then focused on the unique challenges of printed and displayed palms. By examining artifacts like resolution degradation caused by ink bleeding and moiré patterns, their approach introduced a novel way of detecting forgery through visual and structural anomalies.  

Wang \textit{et al}.~\cite{wang2023anti} broadened the scope with a cutting-edge dual-wavelength system. Their solution captured a more comprehensive range of palm features by integrating static and dynamic biometrics through multispectral imaging. Yao \textit{et al}.~\cite{yao2023palmprint} introduced domain generalization into palmprint anti-spoofing research and constructed a novel, large-scale palmprint attack dataset. This dataset serves as a valuable resource for training and evaluating anti-spoofing algorithms.

Lately, Datwase \textit{et al}.~\cite{datwase2024palmprint} proposed a DL-based approach utilizing a multispectral database to identify spoofed palmprint images. This method leverages the rich information from multiple spectral bands to enhance detection accuracy. Building upon this, Liu \textit{et al}.~\cite{liu2024learning} introduced a domain-adaptive algorithm for cross-domain palmprint anti-spoofing scenarios. By adapting to variations across different domains, this algorithm improves the robustness of spoof detection in diverse environments.

\section{Other Palmprint Recognition Tasks}

\subsection{Cross-Domain Palmprint Recognition}

Cross-domain palmprint recognition tackles the challenge of identifying palmprints accurately across diverse datasets or environments, where variations in factors like lighting, resolution, and imaging devices introduce domain shifts. This field ensures robust and precise recognition by addressing these discrepancies between the enrolled and query palmprint images. Cross-spectral palmprint recognition is a subset of this domain, which focuses on matching images captured under different spectral conditions, such as near-infrared (NIR) and visible light.

Ruan \textit{et al}.~\cite{ruan2024lsfm} proposed the Light Style and Feature Matching (LSFM) method, designed to align features across task-specific layers within a high-dimensional space. This alignment significantly reduces domain discrepancies, thereby enhancing recognition accuracy. Similarly, Shao \textit{et al}.~\cite{shao2019palmgan} introduced PalmGAN, which leverages CycleGAN to generate synthetic images mimicking the target domain, enabling unsupervised recognition through a supervised deep hashing network. Both methods aim to mitigate the challenges posed by domain shifts due to variations in lighting, device differences, and environmental factors. Expanding on these advancements, Shao \textit{et al}.~\cite{shao2019cross} developed a framework leveraging transfer convolutional autoencoders. This method extracts low-dimensional features and iteratively refines feature distributions using a discriminator, effectively bridging domain gaps.

In the specific realm of cross-spectral recognition, Ma \textit{et al}.~\cite{ma2022palmprint} introduced the palmprint translation convolutional neural network (PT-net), which translates NIR images into blue spectrum images to address spectral variance. Similarly, Fei \textit{et al}.~\cite{fei2023learning} proposed a joint learning approach for multispectral palmprint feature extraction. Their method derives spectral-invariant representations, ensuring consistent performance when gallery and probe samples are captured under different spectral conditions.

Zhu \textit{et al}.~\cite{zhu2020cross} addressed cross-spectral domain adaptation by employing low-rank canonical correlation analysis (LRCCA). This technique identifies shared subspaces to capture similarities across spectral data and has demonstrated efficacy in experiments involving 12 cross-spectral recognition tasks. Further contributing to this field, Shao \textit{et al}.~\cite{shao2024generating} investigated cross-dataset palmprint recognition, tackling dataset-specific variations and broadening the applications of cross-domain recognition.

These collective efforts highlight the importance of addressing domain-specific challenges in palmprint recognition, paving the way for developing robust and adaptable palmprint biometric systems.

\subsection{Palm-based Multi-Modality Recognition}
Palmprint recognition methods are often limited to matching within a single modality. However, palm images-including palmprints and palm veins-can be captured using various acquisition devices and under diverse lighting conditions, spanning visible light to near-infrared and far-infrared sources~\cite{jia2021performance}. The heterogeneous nature of features between palmprint and palm vein images presents significant challenges for their effective integration in identity recognition.

Recognition methods leveraging these modalities can be broadly classified into multi-modality fusion, which combines features from multiple modalities, and cross-modal recognition, which focuses on matching between different modalities.

\subsubsection{Multi-modality Fusion Recognition}

When identity is established using a single biometric modality, the system is categorized as an unimodal biometric recognition system. While such systems are straightforward and efficient, they are inherently limited by vulnerabilities, such as susceptibility to forgery and fabricated identities. Palm-based multi-modality fusion systems have emerged as a robust alternative to address these challenges. These systems leverage the complementary features of palmprint and palm vein images—captured and stored during the enrollment phase. During recognition, new inputs are compared against the stored data, enhancing security and reliability through integrating diverse biometric traits.  

Several studies have advanced the palm-based multi-modality fusion, particularly in the fusion of palmprint and palm vein modalities. Wang \textit{et al}.~\cite{wang2008person} proposed an identity recognition algorithm that combines palmprint and palm vein images using Mallat's wavelet as a fusion rule, thereby enriching discriminative information. Similarly, Zhang \textit{et al}.~\cite{zhang2011online} developed a device capable of real-time simultaneous capture of palmprint and palm vein images. Their approach introduced dynamic fusion for adaptive image quality and incorporated a liveness detection algorithm based on image brightness and texture analysis.  

Further innovations include Lin \textit{et al}.~\cite{lin2015combination}, who introduced a method for combined recognition by representing palmprint and palm vein images as grayscale surfaces in three-dimensional space. They achieved accurate identity authentication by calculating the standard deviation of surface differences. Ajay \textit{et al}.~\cite{ajay2017palm} took a different approach, using Harris corner detection to extract regions of interest (ROI) and employing grayscale and Gabor filters for feature extraction. Their system verified identities by computing the Euclidean distance between input images.  

Recent advancements have leveraged DL and multispectral imaging for enhanced performance. Zhao \textit{et al}.~\cite{zhao2019joint} introduced the Joint Deep Convolutional Feature Representation (JDCFR) for hyperspectral palmprint recognition, using a CNN stack to extract features across 53 spectral bands. Wu \textit{et al}.~\cite{wu2021palmprint} proposed a deep hash network for palmprint-palm vein fusion recognition, utilizing spatial and channel-level cascades to improve accuracy.

Building on these efforts, Wang \textit{et al}.~\cite{wang2022multispectral} developed a multispectral recognition framework, integrating features from palmprint and palm vein images through feature-level fusion. Their enhanced CNN model demonstrated superior multimodal recognition capabilities. Most recently, Wu \textit{et al}.~\cite{wu2024fusion} presented a feature-level joint learning method for modality correlation. They achieved effective feature extraction and modality fusion for reliable identity recognition by combining sparse unsupervised projection and partial least squares algorithms.  

\subsubsection{Cross-Modality Recognition}

In practical applications, palmprint samples in gallery and probe sets may be captured under varying environmental and lighting conditions, leading to feature inconsistencies. To address these challenges, recent years have witnessed the development of several methods aimed at improving cross-modal palmprint recognition.

Cho \textit{et al}.~\cite{cho2019extraction} pioneered this area, proposing an RGB-NIR cross-spectral matching system for palmprint and palm vein verification. Their approach matches NIR palm images, containing palmprint and palm vein features, with registered RGB palm images with corresponding features within the system.

Further advancing the field, Su \textit{et al}.~\cite{su2023learning} introduced the Modality-Invariant Binary Feature Learning (MIBFL) method for palmprint-to-palm-vein recognition. This approach projects multimodal palmprint and palm vein images into a high-dimensional space, extracting discriminative features from aligned high-dimensional representations.

Gao \textit{et al}.~\cite{gao2024cross} took a novel approach with their cross-chirality palmprint verification (CCPV) framework. This method allows verification using either hand, requiring only a single stored palmprint template. They also introduced a cross-hand loss function to construct a discriminative and robust feature space for cross-hand matching.

Palm-based multi-modality recognition enhances biometric systems by combining diverse features such as palmprints and palm veins, improving security and reliability beyond unimodal systems. Recent advancements in multi-modality fusion and cross-modality recognition address challenges from varying environmental conditions and feature inconsistencies. Techniques like feature-level fusion and deep learning have significantly improved recognition accuracy and robustness, highlighting the potential for high-security identity verification in palm-based systems.




\subsection{Lightweight Palmprint Recognition}
Existing DL-based methods for palmprint recognition often rely on large models. While these models achieve high recognition accuracy, they come with drawbacks such as prolonged computation times and substantial wastage of computational resources. This issue is particularly pronounced in mobile or edge devices, where resource efficiency is crucial. Researchers have developed lightweight network architectures to address these challenges to reduce computation time and resource consumption~\cite{xu2021effective}.

CNN-Fast, for instance, is designed to accelerate training and inference, making it ideal for real-time applications without compromising performance~\cite{sun2018palmprint}. Similarly, MobileNet-V2, optimized for mobile and embedded devices, offers an efficient solution where computational resources are limited~\cite{michele2019mobilenet}.
  
Zhou \textit{et al}.~\cite{zhou2022improved} introduced an improved lightweight CNN for palmprint recognition based on MobileNetV3. This approach refines the compression factor of the channel attention module, replaces the single-layer activation structure in this module with a multi-layer activation structure, and enhances the functionality of the fully connected layer. These improvements collectively bolster the performance of palmprint recognition systems.  

Similarly, Jia \textit{et al}.~\cite{jia2022eepnet} developed EEPNet, a lightweight Enhanced Efficient Palmprint Network. Based on MobileNetV3, this hybrid architecture combines efficient convolutional layers with attention mechanisms to optimize feature extraction from palmprint images. Designed for embedded systems, EEPNet is particularly for real-time applications on mobile and IoT devices.  

Furthermore, Lin \textit{et al}.~\cite{lin2024unconstrained} proposed a ROI extraction method leveraging a lightweight network. Initially, the YOLOv5-lite network is used for hand detection and preliminary localization. Subsequently, an enhanced version of the lightweight UNet network identifies key points at the valleys between specific fingers. These key points are then used to establish a coordinate system, enabling precise extraction of the final palm ROI.  

Together, these innovative approaches highlight the growing emphasis on lightweight and efficient methods for palmprint recognition, particularly for resource-constrained environments.

\begin{table*}[!t]
\centering
\caption{Comparison of Palmprint Datasets with References (FHP: fixed hand pose, FBHP: flexible hand pose)}
\vspace{-10pt}
\resizebox{.8\textwidth}{!}{%
\begin{tabular}{c|c|c|c|c|c|c}
\hline
\textbf{Dataset} & \textbf{Type} & \textbf{Resolution} & \textbf{Participants} & \textbf{Images per Participant} & \textbf{Acquisition Method} & \textbf{Reference} \\ \hline
\textbf{THUPALMLAB} & Contact & 500 dpi (2040x2040) & 80 & 8 images per hand & FHP &~\cite{dai2010multifeature} \\ \hline
\textbf{LPIDB v1.0} & Contact & 500 dpi (varies) & 51 & 2 (left and right hands) & FHP &~\cite{morales2014lpidb} \\ \hline
\textbf{Bosphorus} & Contact & 120 dpi (varies) & 642 & 3 images per hand & FHP &~\cite{yuksel2011hand} \\ \hline
\textbf{PolyU} & Contact & 72 dpi (128x128) & 193 & 11-27 images per palm & FHP &~\cite{zhang2003online} \\ \hline
\textbf{Multi-Spectral} & Contact & 2500-3000 dpi (varies) & 250 & 12 images per palm & FHP &~\cite{zhang2009online} \\ \hline
\textbf{PV\_790} & Contact & 790 nm NIR (varies) & 209 & 5 samples per session & FHP &~\cite{zhao2020learning} \\ \hline
\textbf{BJTU (V1.0)} & Contact & 72 dpi (292x413) & 173 & 10 samples per palm & FHP &~\cite{mu2011mean} \\ \hline
\textbf{CASIA} & Contactless & 300-600 dpi (varies) & 312 & 8-17 images per hand & FBHP &~\cite{hao2008multispectral} \\ \hline
\textbf{UST} & Contactless & 1280x960 pixels & 287 & 10 images per participant & FBHP &~\cite{kumar2003personal} \\ \hline
\textbf{IITD} & Contactless & 256 dpi (varies) & 230 & 5-6 images per palm & FBHP &~\cite{iitd} \\ \hline
\textbf{GPDS} & Contactless & 120 dpi (256 grayscales) & 100 & 10 right-hand images per participant & FBHP &~\cite{travieso2004optimization} \\ \hline
\textbf{NTU-CP-v1} & Contactless & 420x420 to 1977x1977 pixels & 328 & 7-10 images per palm & FBHP &~\cite{matkowski2019palmprint} \\ \hline
\textbf{Tongji} & Contactless & 1280x960 pixels (varies) & 300 & 10 images per palm & FBHP &~\cite{zhang2017towards} \\ \hline
\textbf{NTU-PI-v1} & Contactless & 1280x960 pixels (varies) & 1093 & 7 samples per participant & FBHP &~\cite{matkowski2019palmprint} \\ \hline
\textbf{REST} & Contactless & 72 dpi (536 × 1250) & 150 & 1945 images & FBHP &~\cite{charfi2016local} \\ \hline
\textbf{XJTU-UP} & Contactless & Smartphone camera (varies) & 200 & $>$30,000 images & FBHP &~\cite{shao2019efficient} \\ \hline
\textbf{NUIG\_Palm1} & Contactless & Smartphone camera (varies) & 81 & 20 images per participant & FBHP &~\cite{ungureanu2017unconstrained} \\ \hline
\end{tabular}
}
\vspace{-15pt}
\label{data}
\end{table*}

\section{Palmprint Datasets}

Over the past decade, the establishment of numerous palmprint recognition datasets has significantly advanced the accuracy and reliability of palmprint recognition systems. These datasets are broadly categorized based on the image acquisition methodologies, falling into two primary types: constrained and unconstrained datasets. The constrained category is divided into contact-based and contactless-based datasets, each offering unique characteristics and applications.  

Within the contact-based datasets, images are classified as either high-resolution or low-resolution. High-resolution images, defined as having a resolution exceeding 400 pixels per inch~\cite{Chen2020improved}, provide detailed visibility of palmar friction ridges and creases, which are crucial for precise recognition. The acquisition of these datasets typically necessitates direct physical interaction between the hand and the capturing device, ensuring a stable and consistent environment~\cite{Liang2022innovative}. Additionally, the acquisition protocols for constrained datasets mandate a fixed hand position, with subjects required to display their hands in a specific pose—fingers straight and separated.  

In contrast, contactless datasets eliminate the need for physical interaction, capturing images with the hand positioned at a distance from the device~\cite{Zhu2023contactless}. This approach enhances user convenience and minimizes hygiene concerns, making it increasingly relevant for modern applications.  
On the other hand, Unconstrained datasets embrace a more flexible acquisition protocol. These datasets allow subjects to assume any hand pose during the image capture, catering to practical, user-friendly system deployments~\cite{Jing2023prototype}. Such flexibility reflects real-world scenarios where strict hand positioning is often impractical or unnecessary.  

Table~\ref{data} provides a detailed overview of the acquisition protocols for various palmprint datasets, highlighting the distinctions between constrained and unconstrained methodologies. These datasets collectively underpin the development of robust palmprint recognition systems, driving innovation and enabling broader adoption in diverse applications.

In this section, we introduce some typical palmprint datasets based on categories of contact-based and contactless-based approaches, respectively.

\subsection{Contact-Based Palmprint Datasets}

Among contact-based datasets, two high-resolution datasets are particularly noteworthy. The first is \textbf{THUPALMLAB}~\cite{dai2010multifeature}, which includes 1280 high-resolution palmprint images collected from 80 subjects, with eight images per hand for each individual. These images are in grayscale, with dimensions of 2040 $\times$ 2040 pixels and 500 dpi resolution. 

The second dataset \textbf{LPIDB v1.0}~\cite{morales2014lpidb} consists of 102 palmprint images from 51 volunteers (28 males and 23 females) and 380 latent palmprint images obtained from various surfaces. These images were scanned at 500 dpi with 8-bit depth.
Other typical low-resolution contact-based datasets include:
\begin{itemize}
    \item \textbf{Bosphorus Hand dataset (Bosphorus)}~\cite{yuksel2011hand}, containing 1926 images from 642 individuals. Left-hand images were captured using a commercial scanner with hands placed flat on the glass platen. Three images were obtained per individual, ensuring fingers were spaced apart.
    \item \textbf{Hong Kong Polytechnic University Palmprint dataset (PolyU)}~\cite{zhang2003online}, which contains 7752 palmprint images from 386 palms of 193 individuals (131 males and 62 females), with each palm contributing 11 to 27 samples. The dataset also includes an ROI at 128 $\times$ 128 pixels.
    \item \textbf{Multispectral Palmprint dataset (Multi-Spectral)}~\cite{zhang2009online}, which consists of four spectral datasets (red, green, blue, and near-infrared) from 250 volunteers (195 males and 55 females). Each individual contributed 12 images per palmprint, totaling 6000 images from 500 unique palms, with participants aged 20 to 60 years.
    \item \textbf{Beijing Jiaotong University Palmprint Dataset (BJTU (V1.0))}~\cite{mu2011mean}, comprising 3460 grayscale images in BMP format from 173 volunteers. These images, with dimensions of 292 $\times$ 413 pixels and a resolution of 72 dpi, were captured using a Fujitsu fi-60F high-speed scanner. Each individual contributed 10 samples from both left and right palms.
\end{itemize}

\subsection{Contactless Palmprint Datasets}

For contactless datasets, several significant collections have been established:
\begin{itemize}
    \item \textbf{CASIA Palmprint Image dataset (CASIA)}~\cite{hao2008multispectral}, consisting of 5500 samples from 312 participants, with 8 to 17 images per hand.
    \item \textbf{UST Hand Image Dataset (UST)}~\cite{kumar2003personal}, developed by the Hong Kong University of Science and Technology, includes 10 images of both hands from 287 individuals. The images were captured using an Olympus C-3020 digital camera (1280 $\times$ 960 pixels) without special lighting.
    \item \textbf{IIT Delhi Touchless Palmprint Dataset (IITD)}~\cite{iitd}, comprising 2600 samples from 230 individuals, with each individual contributing 5 to 6 samples per palm.
    \item \textbf{GPDS Hand Dataset (GPDS)}~\cite{travieso2004optimization}, featuring ten right-hand images from 100 users, captured using a desk scanner with 256 grayscale levels and 120 dpi.
    \item \textbf{Tongji Palmprint dataset (Tongji)}~\cite{zhang2017towards}, which includes 12000 images from 300 individuals over two sessions spaced two months apart. Each session contributed 10 samples per individual.
    \item \textbf{NTU Palmprints from the Internet (NTU-PI-V1)}~\cite{matkowski2019palmprint}, containing 7781 images from 2035 palms of 1093 participants. This dataset was established by collecting contactless palmprint images online.
    \item \textbf{NTU Contactless Palmprint (NTU-CP-V1)}~\cite{matkowski2019palmprint}, comprising 2478 images from 328 participants (predominantly Asian, with some Caucasians and Eurasians). Image dimensions range from 420 $\times$ 420 to 1977 $\times$ 1977 pixels, with a median size of 1373 $\times$ 1373 pixels.
    \item \textbf{REgim Sfax Tunisian Hand dataset (REST)}~\cite{charfi2016local}, containing RGB images of both hands from 150 subjects aged 6 to 70 years. These images were captured using a low-cost digital camera with dimensions of 1536 $\times$ 1250 pixels at 72 dpi.
    \item \textbf{Xian Jiaotong University Unconstrained Palmprint dataset (XJTU-UP)}~\cite{shao2019efficient}, comprising over 30,000 images (200 palms) captured using five different smartphones, making it the largest palmprint dataset available using smartphone cameras.
    \item \textbf{NUIG\_Palm1}~\cite{ungureanu2017unconstrained}, including 1616 images from 81 participants aged 19 to 55. Each individual contributed 20 images using five smartphones.
\end{itemize}

These datasets are a foundation for advancing palmprint recognition research and development, offering diverse resolutions, capture conditions, and demographic representations.

\section{Evaluation}

\subsection{Evaluation Metrics for Palmprint Recognition}
Accuracy~(ACC)~\cite{wayman2005biometric} and Equal Error Rate (EER)~\cite{zhao2020deep} are widely recognized as the primary metrics for evaluating the performance of palmprint recognition systems. These measures play a crucial role in determining system effectiveness and are frequently adopted across the field of biometric research. 

ACC reflects the overall precision of a system in authenticating users, indicating its ability to identify legitimate individuals correctly. This metric is the percentage of genuine instances or positive matches the system verifies.

On the other hand, EER is a benchmark for assessing verification (one-to-one matching) performance. It represents the exact operational point where the false acceptance rate (FAR) and false rejection rate (FRR) are equal~\cite{jain2004introduction}. This equilibrium offers a balanced evaluation of the system, providing insight into its reliability under optimal conditions.

The equation for calculating ACC is as follows:

\begin{equation}
\text{ACC} = \frac{\text{TP+TN}}{\text{TP+FN+FP+TN}} \times 100,
\end{equation}

\begin{equation}
\text{FAR} = \frac{\text{FP}}{\text{FP+TN}} \times 100,
\end{equation}

\begin{equation}
\text{FRR} = \frac{\text{FN}}{\text{TP+FN}} \times 100,
\end{equation}
where True Positive (TP) refers to correctly identified positive samples, True Negative (TN) is used to classify negative samples accurately, and False Positive (FP) is used to classify negative samples incorrectly as positive.

While EER and accuracy offer quick, single-point evaluations, they lack the depth needed to analyze biometric system performance comprehensively. Instead, Detection Error Tradeoff (DET) ~\cite{lozenski2024learned} and Receiver Operating Characteristic~(ROC) ~\cite{rivera2020news} curves provide richer insights by capturing the trade-offs and operational flexibility required in diverse real-world scenarios.  

The DET curve shows how the FRR and FAR change as the threshold varies. The horizontal axis represents the FAR, while the vertical axis represents the FRR. The closer the curve is to the bottom-left corner, the better the classification performance.
By flipping the DET curve vertically, we can obtain the ROC curve, where the axes are reversed compared to the DET curve. The closer the curve is to the top-left corner, the better the classification performance.

The Cumulative Match Characteristic (CMC) curve is a widely used performance evaluation metric in identification (one-to-many matching) tasks. It plots the rank \(k\) on the X-axis and the top-\(k\) accuracy \(\text{top}(T, k)\) on the Y-axis. For a given threshold \(T\), \(\text{top}(T, k)\) values are calculated by varying the rank \(k\) from small to large. These points, \((k, \text{top}(T, k))\), are then connected to generate the CMC curve. A CMC curve closer to \(y = 1\) indicates better identification performance of the algorithm.

\begin{table*}[!t]
\centering
\caption{The performance of different methods.}
\resizebox{.8\textwidth}{!}{%
\begin{tabular}{llp{3cm}p{4cm}ll}
\hline
Method           & Year & Datasets  & Architecture    & Closed-Set Performance (\%)    & Open-Set Performance (\%)      \\ \hline
PalmNet~\cite{genovese2019palmnet}           & 2019 & Tongji, IITD, REST, CASIA & Gabor responses and PCA-based unsupervised method                                                                                            & \begin{tabular}[c]{@{}l@{}}ACC = 99.83, EER = 0.16 (Tongji)\\ ACC = 99.37, EER = 0.52 (IITD)\\ ACC = 97.16, EER = 4.50 (REST)\\ ACC = 99.77, EER = 0.72 (CASIA)\end{tabular}           & -         \\ \hline                                                            
CompNet~\cite{liang2021compnet}                           & 2021 & Tongji, IITD, REST, XJTU-UP & Constrained learnable Gabor filters-based method (containing multisize competitive blocks)                                                                                  & \begin{tabular}[c]{@{}l@{}}ACC = 100, EER = 0.018 (Tongji)\\ EER = 0.628 (IITD)\\ EER = 3.211 (REST)\\ EER = 0.170 (XJTU-UP)\end{tabular}           & -          \\ \hline
EEPNet~\cite{jia2022eepnet}      & 2022 & PolyU, Tongji, Multi-Spectral, NTU-PI-v1, NTU-CP-v1 & MobileNet-V3-based method
& \begin{tabular}[c]{@{}l@{}}ACC = 99.95, EER = 0.0002 (PolyU)\\ ACC = 100, EER = 0.0025 (Tongji)\\ ACC = 100, EER = 0.0010 (Multi-Spectral)\\ ACC = 24.26, EER = 22.9057 (NTU-PI-v1)\\ ACC = 94.56, EER = 4.6230 (NTU-CP-v1)\end{tabular}         & -        \\ \hline
Rs-BFL~\cite{li2022row}     & 2022 & IITD, CASIA & Binary feature learning-based method  with l2,1 norm to make the projection matrix more discriminative         & \begin{tabular}[c]{@{}l@{}}ACC = 99.6957$\pm$0.1604 (IITD)\\ ACC = 97.9538$\pm$1.8095 (CASIA)\end{tabular}      & ACC = 98.8043$\pm$0.6276 (IITD)                                              \\ \hline
DC\_MDPR~\cite{zhao2022double}    & 2022 & Multi-Spectral, Tongji, IITD, GPDS, CASIA & multiview palmprint representation (with double-cohesion strategy)       & \begin{tabular}[c]{@{}l@{}}ACC = 100 (Multi-Spectral)\\ ACC = 99.44$\pm$0.12 (Tongji)\\ ACC = 98.43$\pm$0.22  (IITD)\\ ACC = 98.93$\pm$0.58  (GPDS)\\ ACC = 99.11$\pm$0.90 (CASIA)\end{tabular}     & -      \\ \hline
BEST~\cite{yulin2023best}     & 2023 & IITD, CASIA, PolyU & Approximate nearest neighbor-based method & \begin{tabular}[c]{@{}l@{}}ACC = 98.09, EER = 1.38 (IITD)\\ EER = 0.25 (CASIA)\end{tabular}      & ACC = 98.48, EER = 0.69 (PolyU)   \\ \hline
AMGNet~\cite{fan2023amgnet}        & 2023 & PolyU, IITD, CASIA, NTU-CP-v1 & Multilevel Gabor filter-based method (including principal line and wrinkle Gabor convolution modules)                                           & \begin{tabular}[c]{@{}l@{}}ACC = 99.713, EER = 0.2865 (PolyU)\\ ACC = 99.754, EER = 0.1638 (IITD)\\ ACC = 99.206, EER = 0.7613 (CASIA)\\ ACC = 95.959, EER = 2.6655 (NTU-CP-v1)\end{tabular}    & -     \\ \hline
 CCNet~\cite{yang2023comprehensive} & 2023 & PolyU, Tongji, IITD, Multi-Spectral & Learnable Gaber filter-based method (including three branches and the comprehensive competition mechanism)               & \begin{tabular}[c]{@{}l@{}}ACC = 100, EER = 0.00006 (PolyU)\\ ACC = 100, EER = 0.00004 (Tongji)\\ ACC = 100, EER = 0.18 (IITD)\\ ACC = 100, EER = 0 (Multi-Spectral)\end{tabular}                        & \begin{tabular}[c]{@{}l@{}}ACC = 99.76, EER = 1.58 (Tongji-\textgreater{}PolyU)\\ ACC = 99.63, EER = 2.01 (IITD-\textgreater{}PolyU)\\ ACC = 98.48, EER = 2.22 (PolyU-\textgreater{}Tongji)\\ ACC = 98.35, EER = 3.22 (IITD-\textgreater{}Tongji)\end{tabular} \\ \hline
FFLNet~\cite{fei2024learning}      & 2024 &Multi-Spectral, CASIA & Fourier-based feature learning network (with frequency-aware feature learning module)               & \begin{tabular}[c]{@{}l@{}}ACC (Multi-Spectral): \\ 99.63$\pm$0.06 (Blue-NIR), 99.45$\pm$0.09 (NIR-Blue), \\ 99.70$\pm$0.06 (Green-NIR), 99.47$\pm$0.14 (NIR-Green), \\ 99.98$\pm$0.07 (Red-NIR), 99.94$\pm$0.01(NIR-Red)\\ ACC (CASIA): \\ 86.13$\pm$0.53 (460-850nn), 80.96$\pm$0.54 (850-460nn), \\ 94.03$\pm$0.60 (630-850nn), 93.66$\pm$0.34 (850-630nn), \\ 93.46$\pm$0.90 (700-940nn), 93.50$\pm$0.49(940-700nn)\end{tabular} & \begin{tabular}[c]{@{}l@{}}ACC (CASIA): \\ 57.00$\pm$3.50 (460-850nn), \\57.20$\pm$3.39 (850-460nn), \\ 80.00$\pm$1.36 (630-850nn), \\79.50$\pm$1.87 (850-630nn), \\ 82.25$\pm$2.78 (700-940nn), \\80.75$\pm$2.44(940-700nn)\end{tabular}                                              \\ \hline
PDFG~\cite{shao2024learning}         & 2024 & XJTU-UP, CASIA& Fourier-based data augmentation method (with data-level and feature-level generalization)       & -    & \begin{tabular}[c]{@{}l@{}}ACC = 88.95, EER = 2.97 (XJTU-UP)\\ ACC = 92.82, EER = 4.07 (CASIA)\end{tabular}                                                          \\ \hline
\end{tabular}
}
\vspace{-20pt}
\label{methods}
\end{table*}

\subsection{Performance Comparisons}

To comprehensively explore recent advancements in palmprint recognition, this paper systematically reviews notable deep learning-based methods, as summarized in Table~\ref{methods}.

The analysis reveals a dichotomy in performance: while many methods achieve near-perfect results in closed-set settings, their effectiveness in open-set scenarios remains a significant challenge. For instance, CCNet and AMGNet report exceptional closed-set performance, with ACC reaching 100\% on multiple datasets, such as Tongji and PolyU. Yet, their open-set metrics highlight room for improvement, with EERs of 1.58\% and 1.68\%, respectively. Similarly, Rs-BFL~\cite{li2022row} demonstrates competitive closed-set accuracy on ITTD and CASIA but encounters difficulty generalizing to unseen users, as evidenced by its open-set ACC of 98.80\% on ITTD.

The trend also underscores the growing adoption of specialized architectures, including learnable Gabor filters (e.g., AMGNet~\cite{fan2023amgnet} and CCNet~\cite{yang2023comprehensive}) and feature-learning techniques like Rs-BFL, which integrate discriminative projection matrices. These innovations enable robust closed-set recognition but struggle with cross-dataset variability, a hallmark of open-set recognition challenges. Furthermore, emerging approaches, such as Fourier-based methods (e.g., PDFG~\cite{shao2024learning}), highlight promising directions for feature-level generalization, with EERs as low as 2.97\% on XJTU-UP.

\section{Challenges and Outlook}
Palmprint recognition has achieved significant advancements in recent years. However, several challenges persist:

\subsubsection{\textbf{Beyond Closed-Set Recognition}}

While significant progress has been made in closed-set recognition, achieving robust generalization in open-set and cross-domain scenarios remains a critical challenge. Bridging this gap will demand innovative strategies emphasizing cross-dataset consistency, domain adaptability, and user-independent performance.

A promising avenue lies in leveraging domain adaptation and generalization techniques~\cite{wang2022generalizing}. These methods equip models to handle domain shifts and perform effectively in previously unseen scenarios. Complementarily, self-supervised learning~\cite{gui2024survey} offers a powerful approach to extracting meaningful representations from unannotated or cross-domain data, facilitating adaptation to new datasets without requiring extensive manual labeling.

The establishment of cross-dataset benchmarks is equally important. Standardized evaluations can illuminate generalization shortcomings, offering insights to enhance model robustness against variations in demographics, sensors, and environmental conditions. 

Techniques like few-shot and zero-shot learning~\cite{song2023comprehensive} also present exciting possibilities, enabling the recognition of unseen palmprints with minimal or no additional data. Together, these strategies promise to advance the field toward more adaptable and inclusive recognition systems.

\subsubsection{\textbf{Data Caveats}} 
Although numerous palmprint benchmark datasets exist for evaluation and comparison, they fall short compared to the extensive resources available for more mature biometrics like face, iris, and fingerprint recognition. These datasets' limited scale and diversity hinder models from generalizing effectively across varied demographics, environments, and imaging conditions—key requirements for robust open-set and cross-domain recognition. Furthermore, the absence of detailed metadata in many datasets restricts opportunities to enhance model performance and interoperability.

Creating comprehensive palmprint datasets requires addressing critical biases to ensure fairness and accuracy. Demographic imbalances—such as underrepresenting certain age groups, genders, or ethnicities—can reduce performance for marginalized groups. Environmental factors, including lighting conditions, scanner angles, and device-specific sensor characteristics, can introduce inconsistencies. Additionally, limited variability in hand poses, orientations, and age-related palmprint changes diminishes model robustness. 

Temporal biases and annotation errors further compound these challenges. To overcome these issues, datasets must encompass diverse demographics, environmental settings, device types, poses, and time intervals. Balanced sampling and meticulous labeling are essential for building reliable, generalizable systems.

An increasingly promising solution is synthetic data generation using generative models~\cite{bauer2024comprehensive}. These models can simulate various conditions, such as varying hand poses, environmental factors, and sensor characteristics. By augmenting training data, synthetic generation enhances model robustness and adaptability, paving the way for more comprehensive and effective palmprint recognition systems.

\subsubsection{\textbf{Security and Privacy Frontier}}

Like other biometric systems, palmprint recognition is exposed to various security risks, including spoofing, template compromise, and unauthorized access to sensitive data. The advent of DL-based systems, while enhancing recognition accuracy, introduces new vulnerabilities. For instance, DL models are susceptible to template reconstruction attacks, where adversaries exploit stored biometric templates to reverse-engineer and recreate palmprint images. This poses a significant threat, as reconstructed images can be used for unauthorized access.

Moreover, DL models face adversarial attacks, where imperceptible perturbations added to input images can significantly degrade the system's recognition performance or even manipulate the model's output~\cite{zhou2022adversarial}. These attacks exploit the sensitivity of DL models to small input variations, highlighting a critical security gap in palmprint recognition systems.

Addressing these risks requires robust measures such as encryption techniques like homomorphic encryption for secure data handling, template protection strategies like cancelable biometrics to prevent reconstruction and enable revocation, and advanced anti-spoofing methods incorporating multispectral imaging and domain adaptation.
  
\subsubsection{\textbf{LLMs in Biometrics}} 
An emerging direction is the adaptation of large language models (LLMs) through multimodal fusion and novel feature representation techniques~\cite{caffagni2024r}. While LLMs are traditionally designed for natural language processing, their potential to interpret and augment biometric data, including palmprints, is a promising research frontier. Integrating LLMs with image-based models makes it possible to explore multimodal approaches where textual data—such as metadata or contextual information—could be combined with palmprint images to improve recognition accuracy and adaptability. 

This fusion could also lead to innovative ways of representing palmprint features, allowing for richer, more nuanced representations that capture complex patterns and relationships in the data. Integrating LLMs with palmprint recognition systems could open up new pathways for achieving higher accuracy, efficiency, and scalability in biometric security applications.
       
\section{Conclusion}
Palmprint recognition has established itself as a highly promising biometric modality, propelled by innovations in both traditional methods and deep learning (DL)-based approaches. This survey spotlights the strides made in preprocessing, feature extraction, and tackling security and privacy challenges while showcasing DL's transformative role in enhancing accuracy, adaptability, and scalability. However, critical challenges persist, including open-set and cross-domain recognition, safeguarding security and privacy, and addressing dataset limitations. Palmprint recognition is poised to remain a cornerstone of advancing biometric technologies by overcoming these hurdles through innovative solutions.




%





\ifCLASSOPTIONcaptionsoff
  \newpage
\fi





\bibliographystyle{IEEEtran}
\bibliography{Bibliography}
%


\begin{IEEEbiography}[{\includegraphics[width=1in,height=1.25in,clip,keepaspectratio]{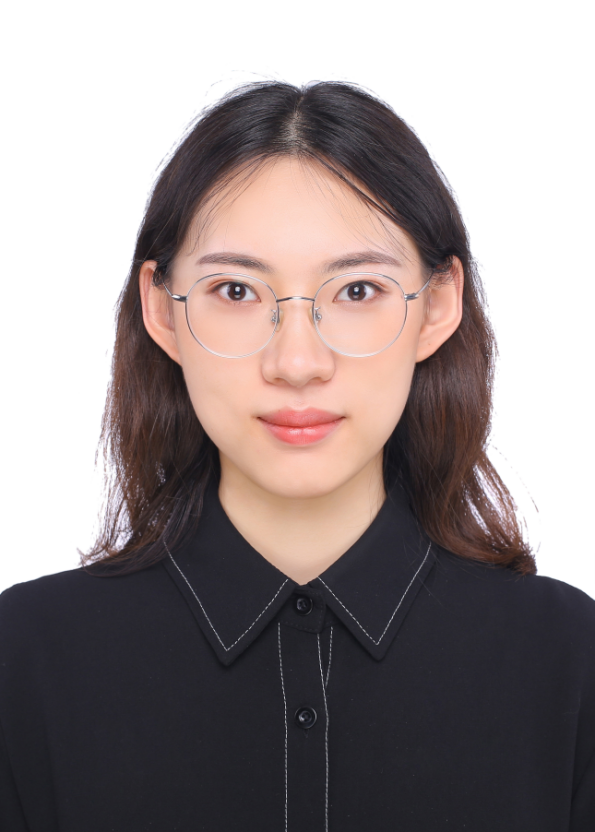}}]{Chengrui Gao}
received the M.S. degree from the School of Electronic Information, Sichuan University, Chengdu, China, in 2021. She is currently pursuing her Ph.D. degree at the College of Computer Science, Sichuan University. She was a joint Ph.D. student with Yonsei University supported by the China Scholarship Council Program in 2023. Her interests include biometrics and pattern recognition.
\end{IEEEbiography}

\begin{IEEEbiography}[{\includegraphics[width=1in,height=1.25in,clip,keepaspectratio]{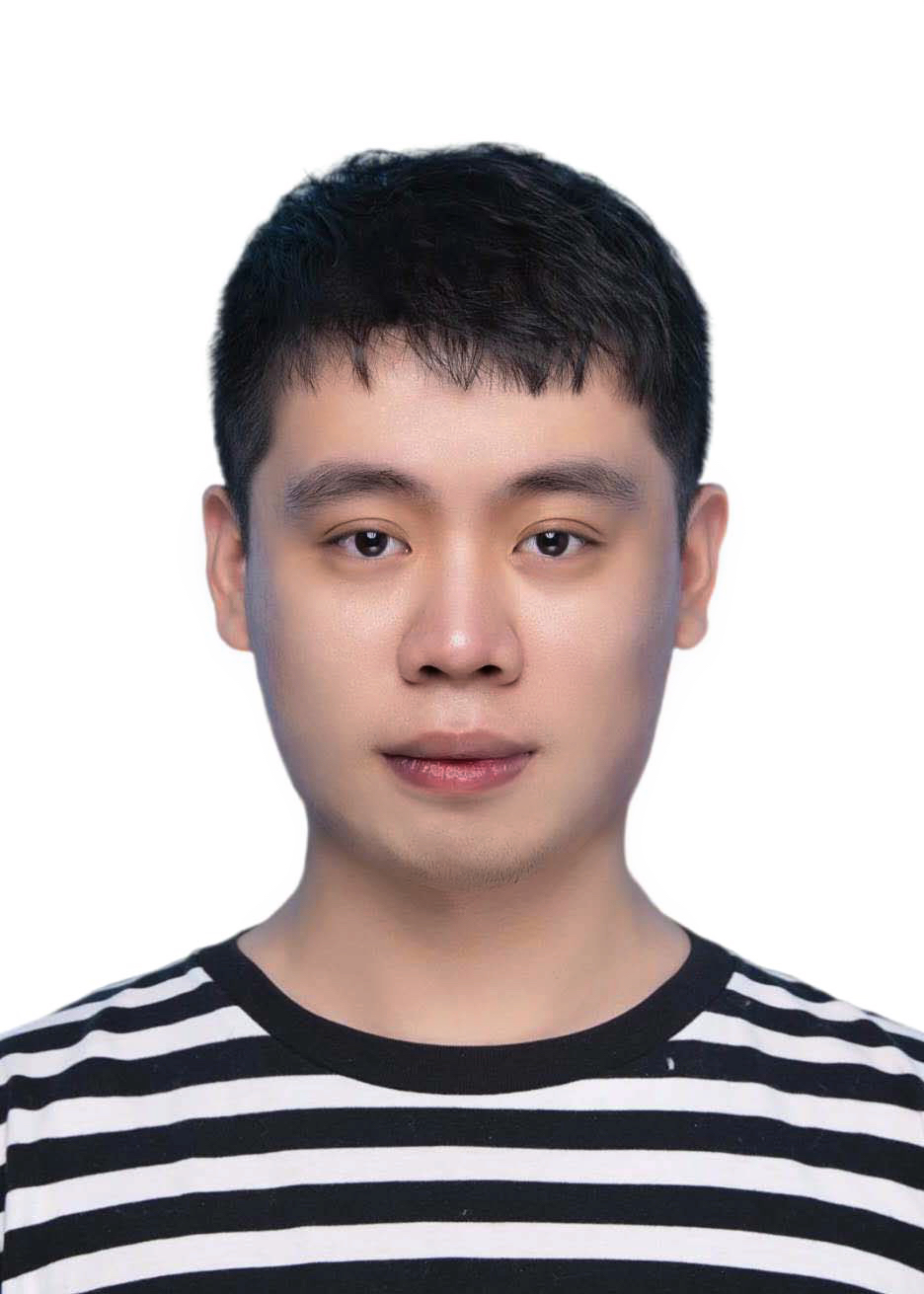}}]{Ziyuan Yang}
received an M.S. degree in computer science from the  School of Information Engineering, Nanchang University, Nanchang, China, in 2021. He is currently pursuing his Ph.D. degree with the College of Computer Science, Sichuan University, China. He is a visiting Ph.D. student at Centre for Frontier AI Research, Agency for Science, Technology and Research (A*STAR), Singapore. His research interests include biometrics, distributed learning, and security analysis.
\end{IEEEbiography}

\begin{IEEEbiography}[{\includegraphics[width=1in,height=1.25in,clip,keepaspectratio]{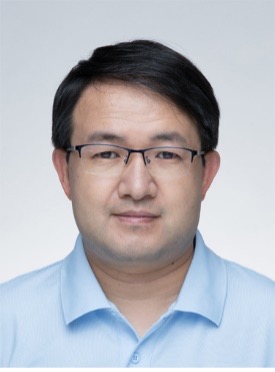}}]{Wei Jia} (Member, IEEE) received the B.Sc. degree in informatics from Central China Normal University, Wuhan, China, in 1998, the M.Sc. degree in computer science from the Hefei University of Technology, Hefei, China, in 2004, and the Ph.D. degree in pattern recognition and intelligence system from the University of Science and Technology of China, Hefei, in 2008., From 2008 to 2016, he was a Research Assistant and Associate Professor with the Hefei Institutes of Physical Science, Chinese Academy of Science. He is currently a Full Professor with the School of Computer Science and Information Engineering, Hefei University of Technology. His research interests include computer vision, biometrics, pattern recognition, image processing, and machine learning.
\end{IEEEbiography}

\begin{IEEEbiography}[{\includegraphics[width=1in,height=1.25in,clip,keepaspectratio]{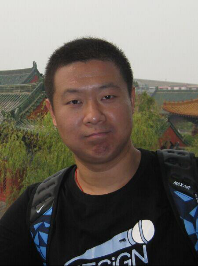}}]{Lu Leng} received his Ph.D degree from Southwest Jiaotong University, Chengdu, P. R. China, in 2012. He performed his postdoctoral research at Yonsei University, Seoul, South Korea, and Nanjing University of Aeronautics and Astronautics, Nanjing, P. R. China. He was a visiting scholar at West Virginia University, USA, and Yonsei University, South Korea. Currently, he is a full professor and the dean of Institute of Computer Vision at Nanchang Hangkong University.
Prof. Leng has published more than 100 international journal and conference papers. He has been granted several scholarships and funding projects, including six projects supported by National Natural Science Foundation of China (NSFC). He serves as a reviewer of more than 100 international journals and conferences. His research interests include computer vision, biometric template protection, biometric recognition, medical image processing, data hiding, etc.
\end{IEEEbiography}

\begin{IEEEbiography}[{\includegraphics[width=1in,height=1.25in,clip,keepaspectratio]{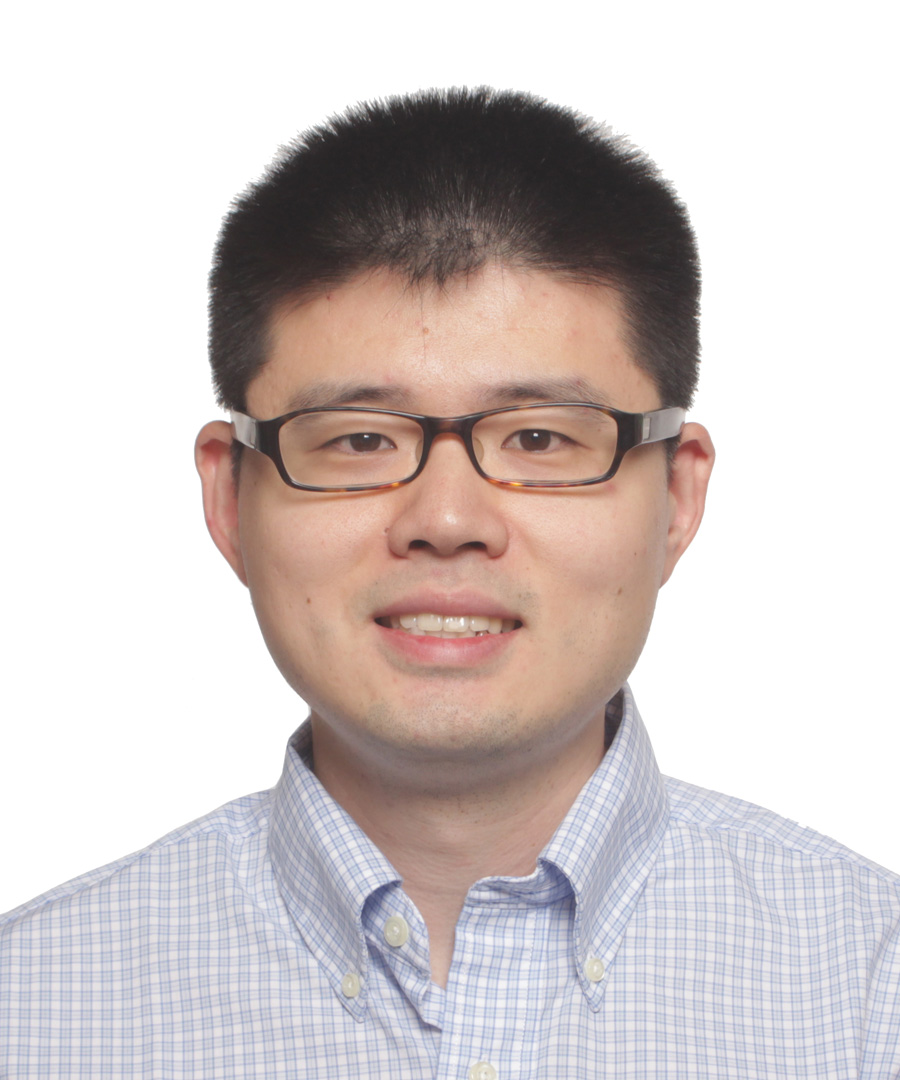}}]{Bob Zhang}
(Senior Member, IEEE) received the B.A. degree in computer science from York University, Toronto, ON, Canada, in 2006, the M.A.Sc. degree in information systems security from Concordia University, Montreal, QC, Canada, in 2007, and the Ph.D. in electrical and computer engineering from the University of Waterloo, Waterloo, ON, Canada, in 2011. After graduating from the University of Waterloo, he remained with the Center for Pattern Recognition and Machine Intelligence, and later, he was a Postdoctoral Researcher with the Department of Electrical and Computer Engineering, Carnegie Mellon University, Pittsburgh, PA, USA. He is currently an Associate Professor with the Department of Computer and Information Science, University of Macau, Macau. His research interests focus on biometrics, pattern recognition, and image processing. Dr. Zhang is a Technical Committee Member of the IEEE Systems, Man, and Cybernetics Society and an Associate Editor of IEEE TRANSACTIONS ON NEURAL NETWORKS AND LEARNING SYSTEMS, Artificial Intelligence Review, and IET Computer Vision.\end{IEEEbiography}

\begin{IEEEbiography}[{\includegraphics[width=1in,height=1.25in,clip,keepaspectratio]{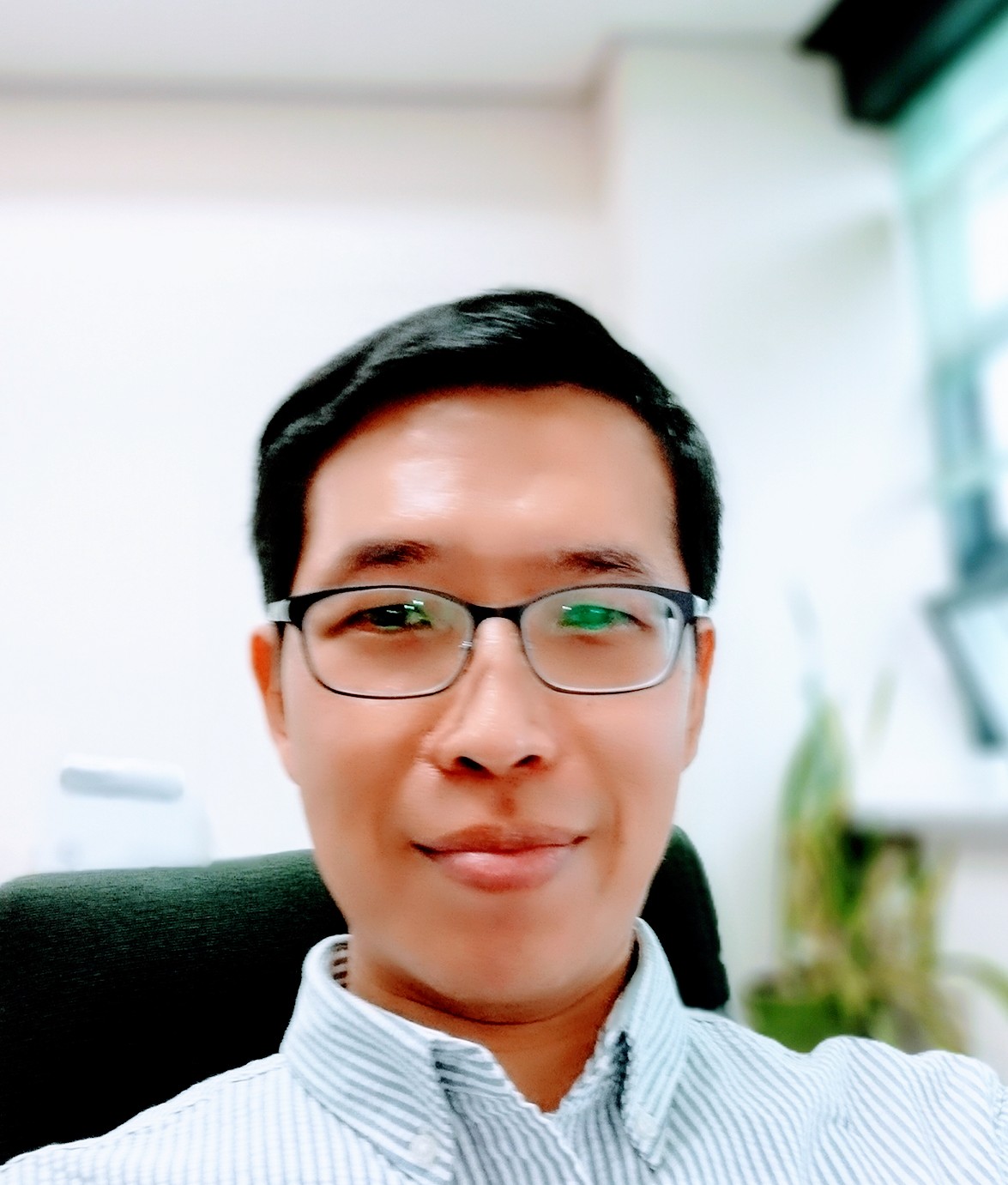}}]{Andrew Beng Jin Teoh}
(Senior Member, IEEE) received the B.Eng. degree in electronic and the Ph.D. degree from the National University of Malaysia in 1999 and 2003, respectively. He is currently a Full Professor with the Electrical and Electronic Engineering Department, College of Engineering, Yonsei University, South Korea. He has published over 350 international refereed journal articles and conference papers and edited several book chapters and volumes. His research, for which he has received funding, focuses on biometric applications and biometric security. His current research interests include machine learning and information security. He was the Guest Editor of the IEEE Signal Processing Magazine and a Senior Associate Editor of IEEE Transactions on Information Forensics and Security, IEEE Biometrics Compendium, and Machine Learning with Applications.
\end{IEEEbiography}

\vfill


\end{document}